**TITLE:**
Explainable Biomedical Hypothesis Generation via Retrieval Augmented Generation enabled Large Language Models

**AUTHORS AND AFFILIATIONS:**
Alexander R. Pelletier[1,2,3,*], Joseph Ramirez[1], Irsyad Adam[1,4], Simha Sankar[1], Yu Yan[1,3,4], Ding Wang[1,3,5], Dylan Steinecke[1,3,4], Wei Wang[1,2,3], Peipei Ping[1,2,3,4,5,*]

1. Department of Physiology, UCLA School of Medicine, Suite 1-609, MRL Building, 675 Charles E. Young Dr., LA, CA 90095-1760, USA.
2. Scalable Analytics Institute (ScAi) at Department of Computer Science, UCLA School of Engineering, Los Angeles, CA 90095, USA.
3. NIH BRIDGE2AI Center at UCLA & NHLBI Integrated Cardiovascular Data Science Training Program at UCLA, Suite 1-609, MRL Building, 675 Charles E. Young Dr. South, Los Angeles, CA 90095-1760, USA.
4. Medical Informatics, University of California at Los Angeles (UCLA), CA 90095, USA
5. Department of Medicine (Cardiology), UCLA School of Medicine, Suite 1-609, MRL Building, 675 Charles E. Young Dr. South, Los Angeles, CA 90095-1760, USA.

* Corresponding authors

**SUMMARY:**
RUGGED (Retrieval Under Graph-Guided Explainable disease Distinction) is a computational workflow that integrates Large Language Model (LLM) inference with Retrieval Augmented Generation (RAG) drawing evidence from trustworthy and curated biomedical knowledge bases as well as peer reviewed biomedical text publications. This approach streamlines the identification of explainable and actionable predictions, synthesizing new knowledge from up-to-date information, to pinpoint promising directions for hypothesis-driven investigations.

**ABSTRACT:**
The vast amount of biomedical information available today presents a significant challenge for investigators seeking to digest, process, and understand these findings effectively. Large Language Models (LLMs) have emerged as powerful tools to navigate this complex and challenging data landscape. However, LLMs may lead to hallucinatory responses, making Retrieval Augmented Generation (RAG) crucial for achieving accurate information. In this protocol, we present RUGGED (Retrieval Under Graph-Guided Explainable disease Distinction), a comprehensive workflow designed to support investigators with knowledge integration and hypothesis generation, identifying validated paths forward. Relevant biomedical information from publications and knowledge bases are reviewed, integrated, and extracted via text-mining association analysis and explainable graph prediction models on disease nodes, forecasting potential links among drugs and diseases. These analyses, along with biomedical texts, are integrated into a framework that facilitates user-directed mechanism elucidation as well as hypothesis exploration through RAG-enabled LLMs. A clinical use-case demonstrates RUGGED's ability to evaluate and recommend therapeutics for Arrhythmogenic

Cardiomyopathy (ACM) and Dilated Cardiomyopathy (DCM), analyzing prescribed drugs for molecular interactions and unexplored uses. The platform minimizes LLM hallucinations, offers actionable insights, and improves the investigation of novel therapeutics.

**INTRODUCTION:**
Hypothesis generation process in biomedical research is essential to uncover novel protein-disease associations to understand pathogenesis and unlock therapeutic potential. This process draws evidence from existing biomedical knowledge, synthesizing findings based on logic leads embedded within peer-reviewed literature, such as the 36 million publications comprising PubMed, and integrating high-confidence curated evidence rooted among the many biomedical knowledge bases. Recent advancements streamline this laborious manual effort by applying text mining on literature corpra[1–3] as well as employ graph-based analyses[4–7] to synthesize relevant information and uncover new avenues for investigation. Despite efforts, our current approaches often lack deep contextual understanding of these fragmented data with limited ability to draw inferences and interactively explore new hypotheses.

Large Language Models (LLMs) shed new light to these challenges, demonstrating deep contextual understanding by training on vast amounts of information across multiple disciplines [8–10]. In the biomedical domain, LLMs have been applied to tasks such as extracting patient information[11], general clinical question answering[12, 13], domain-specific question answering[14], and use in primary clinical care[15]. These models demonstrate ability to reason and draw inferences from complex datasets, making them well-suited for generating hypotheses in biomedical research. Furthermore, some models feature chat-like interaction which engage users and enable dynamic exploration of topics, surpassing the conventional boundaries of knowledge bases and traditional web search engines[16, 17].

Despite their potential, LLMs face challenges, such as hallucinating information, displaying unwarranted confidence in potentially inaccurate explanations, lacking interpretability, and being susceptible to biased or inappropriate content[18–21]. When LLMs are applied directly to hypothesis generation or guiding clinical decision-making, the responses and predictions have high stakes, any errors may potentially guide costly laboratory experiments or influence decisions that affect patient health trajectories[22, 23]. Thus, reliable and trustworthy LLM responses are paramount, as their advice must be firmly rooted in evidence, explicitly laying out their reasoning and substantiating their claims. In these scenarios, interpretability is not a luxury but a necessity for understanding why these models make the predictions they do.

To this end, Retrieval-Augmented Generation (RAG) is a system designed to minimize LLM hallucinations. By identifying and incorporating relevant text documents from reliable and trustworthy sources, RAG grounds LLM responses in evidence, enhancing their accuracy and reliability[24, 25]. For example, integrating an LLM (e.g., ChatGPT) with PubMed allows for the identification of relevant citations to user queries [26, 27]. This method leverages Named Entity Recognition (NER) to connect literature but does not yet integrate information from biomedical knowledge bases or from predictive analyses.

Knowledge Graphs (KGs) have been applied to LLMs for tasks such as fact-checking, transparent reasoning[28–30], encoding knowledge[31], improving question answering[32], and completing knowledge graphs [33]. By encoding factual information from verified sources, KGs enhance the accuracy, transparency, and reliability of LLM responses. Link prediction techniques within these graphs leverage deep learning to identify previously hidden relationships between proteins and diseases, presenting new opportunities for wet lab investigation[3, 34, 35]. Recent advancements in explainable AI predictions further enhance transparency and interpretability of these link prediction tasks, offering insights to support biomedical hypotheses as a viable avenue for investigation[36–38]. This approach, which provides node-level and edge-level insights, would enable identifying key biomedical entities and relevant subgraphs influencing protein-disease predictions. These advancements ensure that LLM-generated decisions are both accurate and evidence-based, significantly boosting their applicability in biomedical research.

In this protocol, we present RUGGED (Retrieval Under Graph-Guided Explainable disease Distinction) as an accessible and efficient workflow for biomedical hypothesis generation. This workflow protocol leverages the vast resources of biomedical literature and knowledge bases for the extraction of relevant information, enabling query-tailored retrieval processes. We employ explainable artificial intelligence predictions to uncover interpretable and actionable insights from the existing biomedical knowledge, thereby enhancing the transparency and utility of predictive models. The completed workflow streamlines the exploration of knowledge graphs and model predictions via RAG-enabled LLMs, facilitating intuitive and informed interactions for researchers, clinicians, and clinical professionals.

This section lays the groundwork for the protocol, with steps to implement this approach described in the following section. Next, we present representative results of this approach with an example use-case. Finally, implications and discussion of this protocol are discussed.

**PROTOCOL:**
**NOTE:** This protocol has been developed in Python and implemented as a Docker container in Windows. The commands provided are based on the Unix environment, within the Docker container. The software is available at https://github.com/pinglab-utils/RUGGED.

**1.     Install Software**

1.1.     **Install Docker.** Visit the Docker website (https://www.docker.com/), click on 'Get started', and choose the appropriate version for your operating system (Windows, macOS, or Linux). Download and run the installer. Verify installation by typing `docker --version` in the terminal; successful installation will report the version of Docker installed.

1.1.1.     **Initialize Inter-container Networking.** Enable Docker containers to be configured to connect to other services on your device (e.g., other Docker containers). Type into the terminal the following command: `docker network create rugged_network`.

1.2.     **Install Git.** Visit the Git website (https://www.git-scm.com/), click on 'Downloads', and choose the appropriate version for your operating system. Download and run the installer. Verify installation by typing `git --version` in the terminal; successful installation will report the version of Git installed.

1.3. **Enable GPU acceleration.** (Recommended, optional) This step enables GPU acceleration for local LLM and Explainable AI predictive analysis and greatly decreases runtime of the software. If your device has an NVIDIA RTX GPU, install the necessary drivers and the CUDA Toolkit from NVIDIA website (https://developer.nvidia.com/cuda-downloads).

1.4. **Set up Large Language Models (LLMs) services.** RUGGED supports the OpenAI API for models such as GPT-3.5 and ChatGPT-4o, and local models using Ollama (e.g., Llama3). Choose the appropriate service based on your needs.

1.4.1. **Obtain an OpenAI API Key.** If using OpenAI services, no software installation is needed. Proceed to OpenAI's website (https://openai.com/blog/openai-api) to create an account and obtain an API key by loading funds into the account. Save this key in a safe place, as it will be needed for software configuration.

NOTE: OpenAI API is a paid service. At time of publication, OpenAI API costs for GPT4 is $30.00/1M tokens (visit https://openai.com/pricing).

1.4.2. **Install Ollama.** If using a local LLM, install Ollama on your device or download the Docker container. To install Ollama, visit the Ollama website (https://ollama.com/download) and follow installation instructions. To install Ollama on Docker, run the following command:

docker pull ollama/ollama

NOTE: At time of publication, there is no stable release for Ollama on Windows OS. Docker installation is recommended for ease of installation.

1.4.3. **Start Ollama Service.** Ensure the CUDA toolkit from Step 1.3 is installed for optional GPU acceleration and the Ollama Docker image is successfully downloaded or the Ollama GUI is installed. Start the Ollama service.

If using Ollama GUI, run the GUI executable (e.g., ollama.exe)

If using Docker, run:

docker run –name ollama --net rugged_network d -v ollama:/root/.ollama -p 11434:11434 ollama/ollama

If using Docker with GPU acceleration, run:

docker run –name ollama --net rugged_network  -d --gpus=all -v ollama:/root/.ollama -p 11434:11434 ollama/ollama

1.4.4. **Run Ollama.** Determine which Local LLM to use by visiting the full list of Ollama supported models at their website at https://ollama.com/library. Recommended models: llama3, llama2, mistral, mixtral. If using Docker, open the command line and type "docker exec run ollama run MODEL", replacing MODEL with the model name; if using Ollama GUI, type "ollama run MODEL".

1.5. **Set up the Knowledge Graph Database.** RUGGED supports Neo4j, accessible as a Docker container, Neo4j Desktop, or Neo4j AuraDB online server. Choose the appropriate service based on your needs, following the below step for the corresponding service.

1.5.1. **Neo4j Docker.** Run the following commands to set up Neo4j in Docker, with PATH_TO_FOLDER as the full file-path for the folder (e.g., /Users/username/RUGGED). For more details on troubleshooting, refer to the Neo4j Docker website (https://hub.docker.com/_/neo4j).

docker pull neo4j
docker run –name neo4j --net rugged_network --publish=7474:7474 --publish=7687:7687 -d -v PATH_TO_FOLDER\neo4j\data:/data neo4j

NOTE: When using Neo4j in Docker for the first time, initialize the service by setting up the default username and password. Run the neo4j_setup.py script (e.g., python neo4j_setup.py) or log in to the web interface at http://localhost:7474.

1.5.2. **Neo4j Desktop.** If using Neo4j Desktop, download and install it from Neo4j website (https://neo4j.com/) and start the application. Create a new project by clicking "New" and naming it appropriately, then click "Add" to create a new Database Management System (DBMS). Select "Local DBMS", set a password, click "Create", then click "Start" and wait for the service to start. A green "ACTIVE" text indicates the DBMS is running.

1.5.3. **Neo4j AuraDB.** Visit the Neo4j website at (https://neo4j.com/cloud/aura-free/) create an account and log in. Select "New Instance" to create an empty instance and save the URI and initial password to access the bolt interface (e.g., bolt://myurl.neo4j.com). Start the instance by clicking the play button if it has not started already, which will display the connection URIin the instance information box.

NOTE: Neo4j AuraDB offers a free tier up to 200,000 nodes and 400,000 relationships. For larger graphs, visit Neo4j pricing.

1.6. **Set up the RUGGED environment.** Run the following commands in the terminal to download the RUGGED Docker image and code repository:

docker pull pinglabutils/rugged:latest
git clone https://github.com/pinglab-utils/RUGGED

Verify the downloaded Docker images by typing `*docker images*`. All Docker images from the previous step should be listed (e.g., Ollama, Neo4j).

1.6.1. **Configure OpenAI Service.** If using OpenAI service, ensure your account and associated API key from Step 1.4.1 have sufficient funds. Modify RUGGED configuration files by editing the configuration file at 'RUGGED/config/openai_key.txt' and add the API key to the file.

1.6.2. **Configure OpenAI Agents.** Determine which LLM agents within RUGGED's system will use OpenAI models. Modify the configuration file at 'RUGGED/config/llm_agents.json' and update the agent fields to specify the selected OpenAI model version. Determine which OpenAI models to use from OpenAI's model documentation at https://platform.openai.com/docs/models. Recommended models: "gpt-3.5-turbo", "gpt-4o".

1.6.3. **Configure Ollama API Endpoint.** By default, Ollama is accessible via API at 'http://localhost:11434'. If using a different service endpoint, modify the configuration files at 'RUGGED/config/ollama_config.json' and update 'OLLAMA_URI' field accordingly.

1.6.4. **Configure Ollama Agents.** Determine which LLM agents within RUGGED's system will use Ollama. Modify the configuration file at 'RUGGED/config/llm_agents.json' and update the agent fields to specify 'ollama' as the selected model.

1.6.5. **Configure the Neo4j Endpoint.** The default password and username for Neo4j in Docker is initialized in the 'RUGGED/config/neo4j_config.json' configuration file. If following steps 1.5.2 or 1.5.3, update the configuration file with the appropriate username, password, and URI by modifying the 'uri', 'username', and 'password' fields.

1.7. **Start the RUGGED Service.** RUGGED integrates several services, which must be individually tested to ensure the software is working as expected. Start the RUGGED service by running command:

```
docker run -it --net rugged_network –gpus=all -v PATH_TO_FOLDER\RUGGED\:/data ping-lab-utils:RUGGED /bin/bash
```

To verify the services are working as expected, navigate to the RUGGED directory and execute the Steps 1.7.1. through 1.7.4. in this terminal window.

1.7.1. **Verify LLM Service Functionality.** Execute test scripts to verify OpenAI and/or Ollama services are functioning as expected. Navigate to the 'test' folder in the RUGGED directory and execute the following commands, as applicable:

```
cd ./test
python test_openai.py
python test_ollama.py
```

1.7.2. **Verify Named Entity Recognition Service Functionality.** By default, the required packages to execute the Named Entity Recognition of user queries are installed in the RUGGED. Execute a test script to verify it is functioning properly.

```
python test_ner.py
```

1.7.3. **Verify Neo4j Service Functionality.** Execute test scripts to verify the Neo4j service is functioning as expected.

```
python test_neo4j.py
```

1.7.4. **(Optional) Verify HTTP Access to Neo4j.** Open a web browser and visit the Neo4j user interface. For Neo4j in Docker or Desktop, the default URL is http://localhost:7474. For Neo4j AuraDB, use the link from step 1.5.3.

1.8. **(Optional) Troubleshooting.** Ensure the services supporting RUGGED are verified ahead of time to anticipate issues when executing the rest of the protocol. Troubleshoot any unsuccessful tests from Step 1.7., if they exist. Test scripts will provide error messages describing the issues.

1.8.1. **Verify Docker Containers.** Confirm all Docker containers are running by using `docker ps` in the terminal. The following containers should be running:

RUGGED docker container
Neo4j docker container (optional)
Ollama docker container (optional)

1.8.2. **Verify Networking Ports.** For Docker services, ensure the correct ports are open and check logs with "docker logs neo4j" or "docker logs ollama". By default, Neo4j uses ports 7474 for http, 7687 for its bolt interface; Ollama uses port 11434.

1.8.3. **Verify Service Applications.** For applications installed directly on the device (e.g., Ollama and Neo4j Desktop), open the applications to confirm they are running.

1.8.4. **Verify Web Services.** For Neo4j AuraDB, log into the website and verify the service is running.

1.8.5. **Verify Firewall Rules.** Modify device firewall rules to ensure the firewall is not blocking any external services.

1.8.6. **Restart Device.** If issues are not resolved, restart the device and retry from Step 1.8.1.

1.8.7. **Open an Issue.** If problems persist, please open an issue on the RUGGED GitHub (https://github.com/pinglab-utils/RUGGED).

2. **Accessing Biomedical Knowledge and Information Extraction**

NOTE: These steps outline the process to incorporate biomedical knowledge and literature using results from two different information extraction pipelines: 1) the CaseOLAP LIFT biomedical text mining pipeline[3] and 2) the Know2BIO knowledge graph construction workflow[7]. To use RUGGED with your own data, proceed to Step 4.

2.1. **Biomedical Literature Extraction.** CaseOLAP LIFT is a computational protocol designed to investigate sub-cellular proteins and their associations with disease through biomedical literature text mining. This workflow enables the identification of relevant biomedical text and extracting high-level protein-disease relationships, which will inform our RAG workflow, enhancing analyses with targeted insights from biomedical reports.

2.1.1. **Run CaseOLAP LIFT Text Mining Analysis.** Visit the CaseOLAP LIFT JoVE Protocol (https://app.jove.com/t/65084) and complete Steps 1 through 3; Steps 4 and 5 are not necessary for this analysis.

2.1.2. **Move Processed Text Documents.** Successful completion of step 3 of this pipeline will result in a large number of biomedical text documents parsed and downloaded (pubmed.json) along with their full text (pmid2full_text_sections.json) found in the data folder of CaseOLAP LIFT. Move these files into the data folder of RUGGED using the below commands:

mv PATH_TO_FOLDER/caseolap_lift/caseolap_lift_shared_folder/data/pubmed.json PATH_TO_FOLDER/RUGGED/data/text_corpus mv PATH_TO_FOLDER/caseolap_lift/caseolap_lift_shared_folder/data/pmid2full_text_sections.json PATH_TO_FOLDER/RUGGED/data/text_corpus

2.1.3. **Move Text Mining Results.** Successful execution of the text mining pipeline will result in a knowledge graph formatted list of protein-disease associations (merged_edge_list.tsv) found in the result/kg folder. Number of publications and protein-disease associations will vary depending on settings used in Steps 1-3 of the protocol (see **Table 2** for example). Move this file into the data folder of RUGGED using the below commands:

mv PATH_TO_FOLDER/caseolap_lift/caseolap_lift_shared_folder/result/graph_data/merged_edge_list.tsv PATH_TO_FOLDER/RUGGED/data/knowledge_graph

2.2. **Biomedical Knowledge Extraction.** This step details the knowledge graph construction using Know2BIO, a comprehensive biomedical knowledge graph integrating data from 30 diverse sources. It features continuous updates and multi-modal data integration, which will be used to support our Retrieval-Augmented Generation (RAG) workflow.

2.2.1. **Clone Know2BIO Repository.** Clone the repository by typing in the command line, using the below command. Navigate to the Know2BIO repository.

git clone https://github.com/Yijia-Xiao/Know2BIO.git
cd Know2BIO

2.2.2. **Prepare Input Data and Licenses.** Navigate to the dataset folder and follow the instructions in the README.md file. Note that some resources require the creation of user accounts (e.g., UMLS thesaurus, DrugBank).

cd dataset

2.2.3. **Download Knowledge Base Resources.** Execute the create_edge_files.py script and monitor progress of the knowledge graph extraction pipeline. Successful execution will result in a large number of biomedical entities represented in a .csv file in the Know2BIO/dataset/output folder.

python create_edge_files.py

2.2.4. **Construct Knowledge Graph.** Execute the prepare_kgs.py script to integrate the information extracted in the previous step. This script automatically combines the extracted relationships into a unified knowledge graph, formatting the graph by data source and domain.

python prepare_kgs.py

2.2.5. **Verify Output.** Verify the completed kg files in the Know2BIO/dataset/know2bio_dataset directory. The combined knowledge graph will be in the file 'whole_kg.txt' which will be used for the RAG pipeline. The remaining steps in the repository README are not required for this analysis.

2.2.6. **Move Knowledge Graph Results.** Successful execution of this pipeline will result in 'whole_kg.txt' with over 6 million edges, see **Table 3**. Verify the number of lines in the file and move the file into the /data/text_corpus/ of the RUGGED directory using the below command or using the file explorer on your computer.

mv PATH_TO_FOLDER/Know2BIO/dataset/know2bio/whole_kg.txt PATH_TO_FOLDER/RUGGED/data/knowledge_graph

2.3. **Construct a Combined Knowledge Graph.** Steps 2.1 and 2.2 result in a knowledge graph from biomedical resources and high-level protein-disease relationships from text mining, respectively. This step integrates these data within a single unified knowledge graph.

2.3.1. **Verify Results in RUGGED directory.** Verify the knowledge graph construction result file (whole_kg.txt) and the text mining relationship results (merged_edge_list.tsv) are in the knowledge_graph directory within the data folder.

2.3.2. **Iterate Results.** Execute the provided combine_kg_results.py script to combine the text mining results within the knowledge graph results. This script will merge the extracted relationships and entities from both sources into a cohesive knowledge graph, ensuring the data is formatted consistently and integrated correctly. Below is an example command:

python ./rugged/knowledge_graph/combine_kg_results.py --output_file ./data/knowledge_graph/rugged_knowledge_graph.txt ./data/knowledge_graph/whole_kg.txt ./knowledge_graph/merged_edge_list.tsv

2.4. **Filter Knowledge Graph.** (Optional) This step samples a subset of the knowledge graph which will be used for the predictive analysis. This step is crucial for addressing technical limitations, as it filters out irrelevant data and reduces the graph size to execute the deep learning predictive analysis on limited computational resources.

2.4.1. **Identify Relevant Nodes.** Determine which biomedical entities are of interest for the predictive analysis in Step 3. This can be done by reviewing the knowledge graph to pinpoint relevant nodes. For this protocol example, we focus on disease nodes for Arrhythmogenic Cardiomyopathy (ACM) and Dilated Cardiomyopathy (DCM), represented by MeSH_Disease:D002312 and MeSH_Disease:D002311, respectively.

2.4.2. **Filter Knowledge Graph.** Use the filter.py script to tailor the knowledge graph for predictive analysis by focusing on specific subgraphs or diseases of interest. The script employs k-hop filtering from disease nodes of interest to include relevant entities and relationships: Example command below, filtering the graph reachable within 2 nodes from the selected disease nodes.

```
python ./rugged/knowledge_graph/filter.py --k 2 --disease
"MeSH_Disease:D002312,MeSH_Disease:D002311"
--input_file ./data/knowledge_graph/rugged_knowledge_graph.txt --output_file
./data/knowledge_graph/filtered_k2_rugged_knowledge_graph.txt
```

NOTE: Increasing the k-hop value expands the data scope within the graph for prediction analysis but also demands more computational resources. Adjust the k-hop value based on the available resources and the analysis complexity.

**3. Perform Explainable Prediction Analysis.** In this step, we will use GNNExplainer[38] on a Graph Convolutional Network (GCN) to perform prediction analysis. The process involves splitting the filtered knowledge graph into training, validation, and test sets with an 85:5:10 ratio. This split allows us to evaluate the model's performance accurately. The goal is to predict potential edges (relationships) in the knowledge graph, providing insights into previously unknown associations.

3.1. **Ensure the RUGGED Docker container is running.** If the previous terminal window as closed, connect to the Docker container by using the below command:

```
docker exec --it rugged /bin/bash
```

Once connected to the Docker container, navigate to the RUGGED directory.

3.2. **Determine the edge(s) to predict.** Provide the edges as pairs of nodes in a .txt file (e.g., edges_to_predict.txt). The edges already existing in the knowledge graph will be filtered out from the predictions.

3.3. **Run the prediction analysis script**. Specifying the edges to predict and the input knowledge graph as command line arguments. Example command:

```
python ./rugged/predictive_analysis/generate_explainable_prediction.py -p edges_to_predict.txt
-i ./data/knowledge_graph/filtered_k2_rugged_knowledge_graph.txt -o output -n 5 -k 10
```

-p edges_to_predict.txt: Path to the file containing edges to predict.
-i filtered_k2_rugged_knowledge_graph.txt: Path to the input knowledge graph.
-o output: Directory for output files.
-n 5: Number of top predictions to output.
-k 10: Number of top important edges to visualize.

3.4. **Verify results are in the output folder.** Examine the model results in prediction_results.csv and examine the top n predictions within the output folder. Review the top n predictions in the output folder. For each prediction, a graph visualization and relative importance scores for the top k edges are provided. The visualizations illustrate the most pertinent edges contributing to each prediction.

3.5. **Move Predictive Analysis Results.** Once satisfied with the predictive analysis results, move the results into the /data/predictions/ of the RUGGED directory.

NOTE: Explainability AI prediction analysis enhances the interpretability of predictions but requires substantial computational resources and time. Steps 4.1 and 4.2 can be performed concurrently with Step 3.

**4. Hypothesis Generation**

4.1. **Connect to the RUGGED Docker Container.**

4.1.1. **Ensure the RUGGED Docker container is running.** If the previous terminal window was closed, connect to the Docker container by using the below command:

docker exec --it rugged /bin/bash

4.1.2. **Navigate to the RUGGED directory.** The remaining steps will be issued in this command line window and directory.

cd /workspace/RUGGED

4.1.3. **Verify the supporting services are running.** For example, if using Ollama and Neo4j in Docker, ensure the containers are running by typing 'docker ps'. Repeating Step 1.7 to verify services are functioning properly and Step 1.8 to troubleshoot issues if they exist.

4.2. **Prepare RAG Data.** This step sets up the knowledge graph and text corpus for retrieval. Knowledge graph and literature may be substituted with user defined data by following the format outlined in our GitHub repository (https://github.com/pinglab-utils/RUGGED/tree/main/data). Place the data into the data/knowledge_graph/ and data/text_corpus/ directories, respectively.

4.2.1. Ensure the text corpus is in the data/text_corpus/ directory from Step 2.1.6., the knowledge graph with text mining predictions file is located in the data/knowledge_graph/ directory from Step 2.3.3, and the prediction results are in data/predictions/ directory from Step 3.5.

4.2.2. **Populate the Neo4j Graph Database.** Execute the provided script to create the necessary nodes, edges, and node features. This will allow RUGGED to access the data:

python ./neo4j/prepare_neo4j.py

4.2.3. **Index the Text Corpus.** Execute the provided script to index the text corpus. This script will chunk the text into sections of 1000 tokens and create a vector database using the BERT (cite) model to generate embeddings. This process enables RUGGED to retrieve relevant text documents based on user queries:

```
python ./text/prepare_corpus.py
```

4.2.4. **(Optional) Test Neo4j Database Retrieval.** Send a test query to the Neo4j database to ensure it is populated correctly and can return expected results. Verify that the output matches the expected nodes and relationships in the database. Example command:

```
python ./neo4j/test_neo4j_retrieval.py --query "MATCH (n) RETURN n LIMIT 5"
```

4.2.5. **(Optional) Test RAG Corpus Retrieval.** Send a test query to the RAG text corpus to ensure the text retrieval system is working. Check that the retrieved documents are relevant to the query and that the embeddings are functioning as expected. Example command:

```
python ./text/test_rag_retrieval.py --query "example query text"
```

4.3. **Explore Hypotheses with RUGGED.** Start RUGGED in the command line interface to interact with the system and explore your hypotheses. Type the following command in your terminal:

```
python rugged.py
```

4.3.1. **Interact with RUGGED**. Once the CLI is running, you can start querying the system to retrieve relevant information. Use natural language queries or specific commands to interact with the knowledge graph and text corpus. An example use case is presented in Figures 3, 4, and the full transcript is in the Supplementary Materials.

4.3.2. **Query the Knowledge Graph.** To extract specific information from the knowledge graph prepared in Step 2, pose your question in natural language, starting with the keyword "query". For example:

```
query "What example of drug?"
```

4.3.3. **Explore Predictions.** To explore link prediction analyses from Step 3, ask to search for a specific relationship, leading with the keyword "predict". For example:

```
predict "What are some new potential drug targets to treat ACM?"
```

4.3.4. **Literature Retrieval.** To explore documents related to a specific biomedical topic from Step 2, pose the question in natural language, leading with the keyword "search". For example:

```
search "What literature supports the possibility that protein X is related to disease Y?"
```

4.3.5. **Iterate and Refine.** Since RUGGED utilizes the chat-like capabilities of LLM models, user questions can be revised and refined in-context. Based on the initial results, refine your queries and continue exploring different hypotheses. Use the interactive capabilities of RUGGED to iteratively improve your understanding and gather more precise information.

4.3.6. **Rerun Cypher Commands in Neo4j.** (Optional) The knowledge graph query results will provide a Cypher command used to retrieve the information. You can rerun or modify this command by visiting the Neo4j browser interface from Step 1.7.4 (e.g., at http://localhost:7474). Paste and modify the Cypher commands as needed to refine your queries and gather more specific insights from your data.

4.3.7. **Summarize Conversation.** Review the retrieved information and summarize the conversation with RUGGED. Type the keyword "summarize" to output a summary of the interaction to a text file for later analysis. The full text response will be displayed in the terminal.

4.3.8. **Review Chat Logs.** For troubleshooting and reproducibility, full text of the interaction is accessible in the /log/ folder in RUGGED, including intermediate commands and conversations between LLM agents within RUGGED.

4.4. **Shutting Down and Restarting RUGGED.**

4.4.1. **Get Docker Container IDs.** Use docker ps to list all running containers and obtain the container IDs for RUGGED, Neo4j, and Ollama.

docker ps

For all following commands, replace <container_id_for_rugged>, <container_id_for_neo4j>, and <container_id_for_ollama> with the actual container IDs.

4.4.2. **Stop Docker Containers.** Shut down RUGGED and the associated Docker containers using their container IDs.

docker stop <container_id_for_rugged>
docker stop <container_id_for_neo4j>
docker stop <container_id_for_ollama>

NOTE: While shutting down your device will stop all Docker containers automatically, manually stopping these containers first is recommended to prevent potential data loss and ensure all processes close properly.

4.4.3. **Restart Docker Containers.** To restart the RUGGED system, use the container IDs to start the necessary Docker containers.

docker start <container_id_for_rugged>
docker start <container_id_for_neo4j>
docker start <container_id_for_ollama>

4.4.4. **Reattach to Docker Network.** If needed, use these commands to re-attach the containers to the network.

docker network connect rugged_network <container_id_for_rugged>
docker network connect rugged_network <container_id_for_neo4j>
docker network connect rugged_network <container_id_for_ollama>

4.4.5. **Verify Service Functionality.** Upon restart, repeat Step 1.7 and 1.8 to ensure the software is working as expected.

**REPRESENTATIVE RESULTS:**

These representative results were obtained by following the procedure outlined in this protocol. A text mining association analysis was performed following the CaseOLAP LIFT protocol with default parameters, studying 8 broad categories of cardiovascular disease and their association with mitochondrial proteins (GO:0005739)[3]. In total, 635,696 publications through May 2024 were determined as relevant to these diseases; among them, 4,655 high confidence protein-disease associations were identified to inform downstream analyses. A biomedical knowledge graph was constructed using the software code from Know2BIO using default settings in May 2024[7]. The resulting knowledge graph consists of 219,450 nodes and 6,323,257 edges as well as node features for 189,493 nodes, consisting of node descriptions, protein/gene sequences, chemical structure, etc. where available.

The RUGGED system was initialized by constructing vector databases for both the knowledge graph nodes and features as well as the CVD-relevant research publications. All knowledge graph nodes, edges, and node features were chunked using a chunk size of 20 tokens with the BART[39] embedding model to prepare for RAG vector search. Similarly, original contribution and review articles were processed using a chunk size of 500 tokens and the BART embedding model to prepare for RAG vector search. For literature retrieval, full text publications greater than 500 tokens were hierarchically summarized based on the publication subsections by the BART embedding model. The ChatGPT 4o model was used for remaining LLM agents in the system.

An example use-case of hypothesis exploration pertaining to ACM and DCM diseases is outlined in **Figure 3**, with the model response in **Figure 4**. A full transcript of the interaction is in **Supplementary Materials**. To predict novel avenues for therapeutic treatment, the resulting knowledge graph was filtered based on relevance to, including only nodes and edges within 2-hop from Arrhythmogenic Cardiomyopathy (ACM) and Dilated Cardiomyopathy (DCM), represented by MeSH_Disease:D002312 and MeSH_Disease:D002311, respectively. This use-case examines a finer grain investigation of the major cardiovascular disease subtypes, in particular ACM which is a subtype of arrhythmia (ARR). The resulting knowledge graph of 75,621 nodes and 1,376,517 edges was applied to train a graph convolutional neural network prediction model, with evaluation metrics reported in **Table 4**. The top 10 predictions by the model were examined by a graph explainability module, GNNExplainer[38], to identify the top nodes and edges contributing to the prediction.

**FIGURE AND TABLE LEGENDS:**

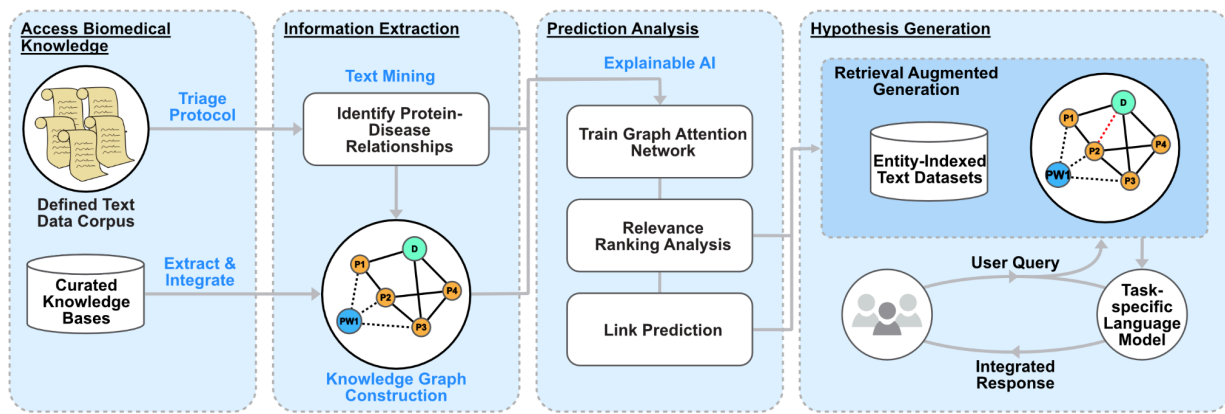

**Figure 1. Retrieval Under Graph-Guided Explainable disease Distinction (RUGGED) Workflow.** RUGGED is comprised of four main components: 1) downloading and processing data from ethically sourced and professionally managed resources (e.g., PubMed and curated biomedical knowledgebases), 2) extracting information from peer-reviewed reports and integrating them into a knowledge graph, 3) identifying explainable predictions regarding the connections among biomedical entities within the knowledge graph, and 4) a Retrieval Augmented Generation (RAG) workflow (detailed in **Figure 2.**) enables a large language model to access validated complex molecular relationships with AI-supported disease predictions to achieve enhanced exploration of biomedical hypotheses.

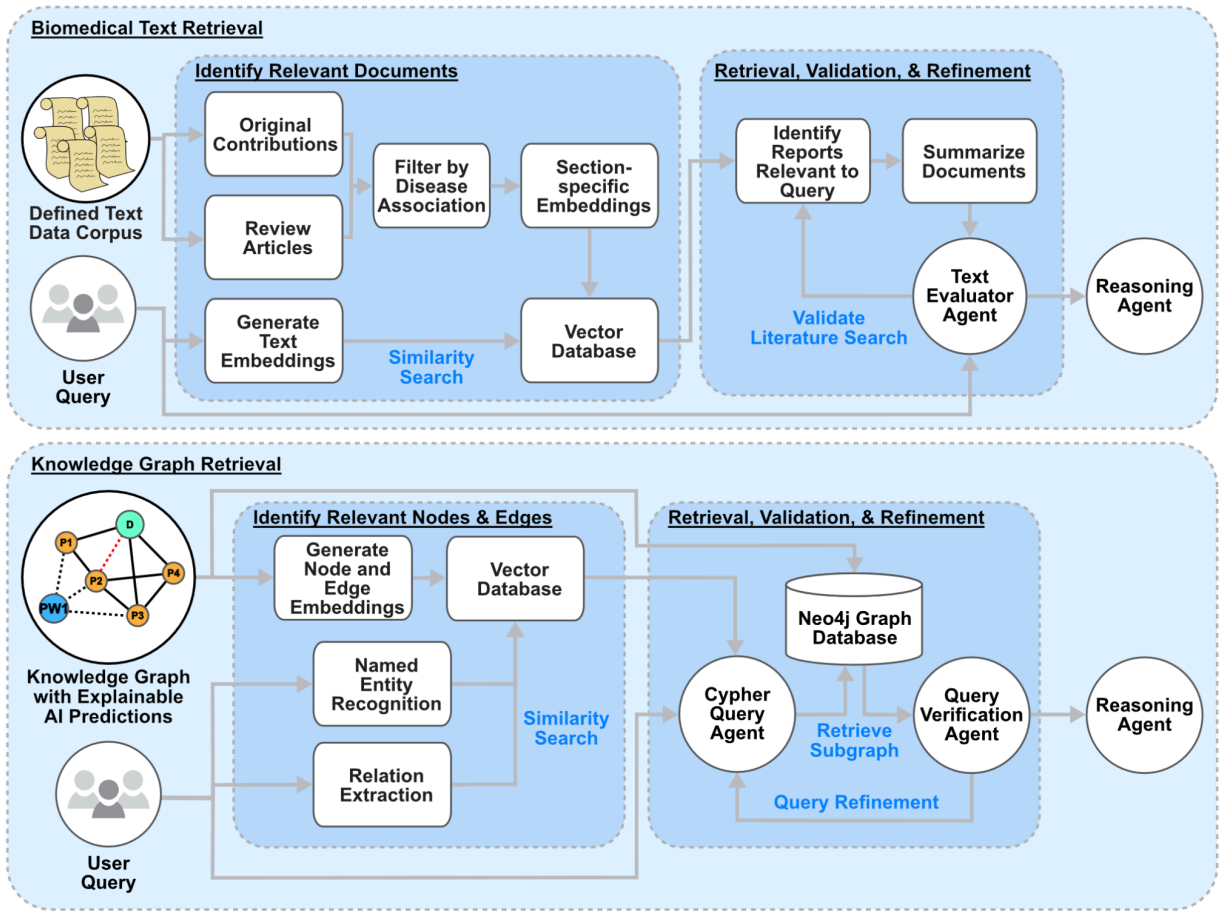

**Figure 2. Retrieval Architecture and Workflow.** The Retrieval Augmented Generation (RAG) framework employs multiple Large Language Model (LLM) Agents, each executing specific roles to access relevant information based on the user query. This system provides citable evidence for the user-facing Reasoning Agent, facilitating user-agent interaction and synthesis of knowledge. 1) Knowledge Graph Retrieval: Named entity recognition and relation extraction connects the user query to relevant entities in the knowledge graph. A similarity search in a vector database identifies pertinent nodes and edges. Data is retrieved from the Neo4j database via Cypher queries generated by the Cypher Query Agent and refined by the Query Verification Agent. 2) Biomedical Text Retrieval: Peer-reviewed original contributions and review articles are filtered based on their relevance to disease associations. A vector database is constructed for each section of the publication, with text evidence weighted according to its context: 70% Abstract, 10% Results, 10% Metadata, and 10% for all other subsections. A similarity search against a text embedding of the user query identifies relevant documents. Summaries are generated using a BERT-based summarizer, with the Text Evaluator Agent refining the search to validate query-document relevance.

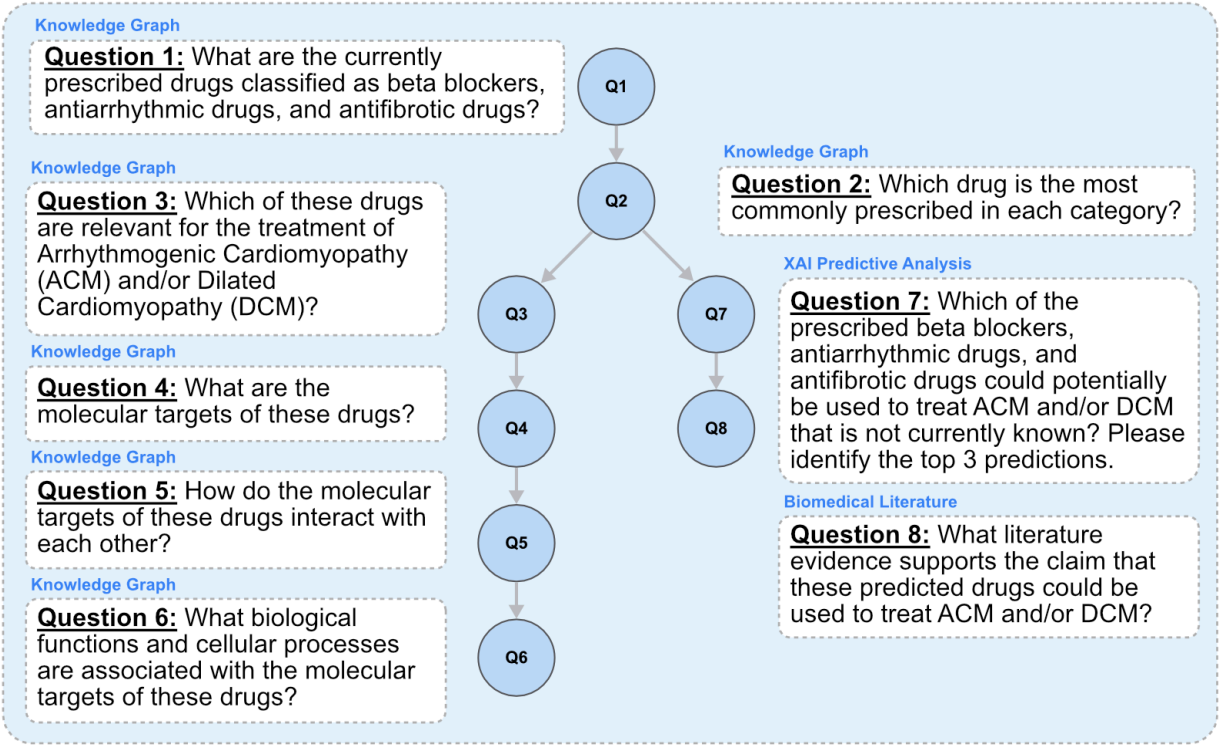

**Figure 3. Hypothesis Exploration Use-case Cascade.** This figure showcases a highlighted use-case focusing on a chain of related biomedical questions and concepts a researcher might pose to the RUGGED system. Queries from the user are proposed to the system in numerical order, with arrows representing logical connections between each question. The system retrieves the necessary relevant information to respond to the query, shown in blue. Examples of system responses are presented in **Figure 4**.

**User:** What are the currently prescribed drugs classified as beta blockers, antiarrhythmic drugs, and antifibrotic drugs?

**RUGGED:** [Querying knowledge graph] Based on the classifications of drugs you provided, our system identified relevant ATC codes for beta blockers, antiarrhythmic drugs, and antifibrotic drugs, which are respectively ATC_Class:C07, ATC_Class:C01B, and ATC_Class:L01X.

Using these classifications, a Cypher query was constructed to extract a list of drugs for each class from a knowledge graph: Beta Blockers (36 drugs), Antiarrhythmic Drugs (29 drugs), Antifibrotic Drugs (53 drugs). [Abbreviated response; see Supplemental for full response]

**User:** Which drug is the most commonly prescribed in each category?

**RUGGED:** [Querying knowledge graph] Using a Cypher query, we analyzed the most connected drug compounds in three categories—Beta Blockers, Antiarrhythmics, and Antifibrotics—from a knowledge graph to identify their frequency and relevance in medical treatments. The drugs were ranked by their connections, highlighting their widespread prescription. Beta Blockers: Pindolol; Antiarrhythmics: Quinidine; Antifibrotics: Sorafenib. [Abbreviated response; see Supplemental for drug descriptions]

**User:** Which of these drugs are relevant for the treatment of Arrhythmogenic Cardiomyopathy (ACM) and/or Dilated Cardiomyopathy (DCM)?

**RUGGED:** [Querying knowledge graph] In response to the inquiry regarding the relevance of specific drugs for treating Arrhythmogenic Cardiomyopathy (ACM) and Dilated Cardiomyopathy (DCM), our system identified relevant MeSH disease matches for ACM and DCM as MeSH_Disease:D002312 and MeSH_Disease:D002311, respectively. We constructed a Cypher query to examine a list of drugs within the categories of Beta Blockers, Antiarrhythmics, and Antifibrotics. The query confirmed the presence of several key drugs for both diseases, providing a list of compounds active in the treatment of these cardiac conditions. Below is a brief description of each of the 7 resulting drug identified in the analysis [Abbreviated response; see Supplemental for drug descriptions]

**User:** What are the molecular targets of these drugs?

**RUGGED:** [Querying knowledge graph] Based on the results of your query, I can provide information about the molecular targets of the specified drugs, which are Beta Blockers and Antiarrhythmics. Here are the details:

Beta Blockers:
1. DB00264 (Metoprolol) has 2 molecular targets: UniProt:P08588 (Beta-1 adrenergic receptor), UniProt:P07550 (Beta-2 adrenergic receptor). [Abbreviated response; see Supplemental for more]

**User:** How do the molecular targets of these drugs interact with each other?

**RUGGED:** [Querying knowledge graph] Based on the results of your query, here's a summary of how the molecular targets of the specified drugs interact with each other:

UniProt:P08588 (Beta-1 adrenergic receptor) interacts with 24 proteins, including Beta-2 adrenergic receptor and several other signaling molecules and receptors. [Abbreviated response; see Supplemental for more]

**User:** What biological functions and cellular processes are associated with the molecular targets of these drugs?

**RUGGED:** [Querying knowledge graph] This analysis examines the molecular targets and biological effects of seven drugs, including five beta blockers (Metoprolol, Propranolol, Nifedipine, Carvedilol, Atenolol) and two antiarrhythmics (Disopyramide, Amiodarone). By analyzing the Gene Ontology (GO) terms associated with the proteins targeted by these drugs, we can elucidate their high-level biological functions and cellular processes. Understanding these interactions is crucial for comprehending how these drugs exert their therapeutic effects. Based on the results of your query, I can provide detailed information about the biological functions and cellular processes associated with the molecular targets of the specified drugs.

**Beta Blockers**
DB00264 (Metoprolol). Metoprolol targets the Beta-1 adrenergic receptor (UniProt:P08588) and the Beta-2 adrenergic receptor (UniProt:P07550). This drug primarily affects heart rate and vascular smooth muscle contraction, leading to decreased heart rate (GO:0001997) and vasodilation (GO:0042310). By blocking these receptors, metoprolol reduces the workload on the heart and helps to manage hypertension and arrhythmias. [Abbreviated response; see Supplemental for more]

**User:** Which of the prescribed beta blockers, antiarrhythmic drugs, and antifibrotic drugs could potentially be used to treat ACM and/or DCM that is not currently known? Please identify the top 3 predictions.

**RUGGED:** [Performing Explainable Link Prediction]
The list of prescribed beta blockers and antiarrhythmic drugs for their known treatments of Arrhythmogenic Cardiomyopathy (ACM) and Dilated Cardiomyopathy (DCM) are explored below. [Showing output for predicted edge between Atenolol (DB00335) and Arrhythmogenic Cardiomyopathy (ACM): Predicted probability of 0.9828]

**Influential Nodes and Paths**
• MeSH_Disease:C566005, MeSH_Disease:C566171): With a score of 0.830, this edge suggests a strong association between these disease nodes, implying potential shared pathways or mechanisms relevant to ACM.
[Edge importances truncated; top 10 importance scores illustrated in graph, right panel]
• ATC_Class:C07BB03, DrugBank_Compound:DB00335): The score of 0.194 highlights the classification and pharmacological context of Atenolol, linking it to its therapeutic use in cardiovascular diseases.
[Continued in right panel]

**Potential Biological Implications**
• Disease Mechanisms: The high-scoring edges between Atenolol and ACM, and other diseases, suggest shared pathogenic pathways. Existing edges between ACM and other diseases include the edge "-diseases_share_genes-" between disease nodes, indicating shared biological mechanisms. Exploring these shared gene associations within the knowledge graph can provide deeper insights into the common pathogenic pathways. This could imply that targeting these shared pathways might have therapeutic effects on ACM due to the underlying common mechanisms.
• ATC Classification: The Anatomical Therapeutic Chemical (ATC) classification system categorizes drugs based on their therapeutic use and chemical characteristics. Atenolol is classified under C07BB03, indicating it is a beta-blocker used in cardiovascular diseases. This classification underscores Atenolol's role in modulating cardiovascular functions, aligning with its potential impact on Arrhythmogenic Cardiomyopathy. Understanding Atenolol's ATC classification helps in recognizing its therapeutic mechanisms and guiding its application in treating ACM.

**Strength and Reliability of the Link Prediction**
• Graph Explainability Metrics: The high scores for the prediction (0.9828) along with high scoring top edges indicate a robust link prediction. These metrics suggest strong associations and relevance of the connections, providing confidence in the predicted link between Atenolol (DB00335) and Arrhythmogenic Cardiomyopathy (D002313).
• Potential for Wet Lab Validation: Given the strong and biologically plausible connections, this link prediction holds promise for experimental validation. The identified pathways and disease interactions can be further investigated in wet lab settings to confirm their roles and therapeutic potential.

**Top10 Important Edges and Predicted Edge Highlighted**

**User:** What literature evidence supports the claim that these predicted drugs could be used to treat ACM and/or DCM?

**RUGGED:** [Performing Literature Retrieval]
The investigation of Atenolol (DB00335) as a potential therapeutic for Arrhythmogenic Cardiomyopathy (ACM) has been supported by a combination of literature evidence. Our system identified four research publications relevant to Atenolol and ACM, including two original research articles and two clinical case reports.

**Original Research Articles**
• Effect of beta-blockers on arrhythmias during six weeks after suspected myocardial infarction (PMID: 387170). This study examined the incidence of arrhythmias in patients with suspected myocardial infarction treated with either propranolol, atenolol, or placebo. The findings indicated that beta-blockers, including atenolol, did not show significant antiarrhythmic action in the dosage used. This study raises doubts about the effectiveness of atenolol in reducing arrhythmias post-myocardial infarction, which could imply limited utility in treating ACM-related arrhythmias.
• Comparison of the effects of xamoterol, atenolol, and propranolol on breathlessness, fatigue, and plasma electrolytes during exercise in healthy volunteers (PMID: 1685993). This research investigated the effects of atenolol and other beta-blockers on heart rate, blood pressure, and electrolyte concentrations during exercise. Atenolol significantly reduced maximum exercise heart rate and blood pressure. The study suggests that atenolol's impact on exercise-induced hyper-kalaemia could be beneficial in managing exercise-induced arrhythmias, which is relevant for ACM.

**Clinical Case Reports**
• Exercise-induced ventricular arrhythmias and sudden cardiac death in a family (PMID: 9106603). This case report describes a family with a history of sudden cardiac deaths and exercise-induced arrhythmias. The patients were treated with atenolol, which significantly reduced exertional arrhythmias as assessed by serial treadmill tests. This suggests atenolol may be effective in preventing exercise-induced arrhythmias, which is pertinent to ACM.
• Sertraline-induced rhabdomyolysis in an elderly patient with dementia and comorbidities (PMID: 19567656). While this report primarily discusses sertraline-induced rhabdomyolysis, it mentions atenolol as part of the patient's medication regimen without adverse interactions. This is less directly relevant but indicates atenolol's general safety profile in complex patient cases.

**Synthesis of Evidence**
The literature provides mixed support for the hypothesis that atenolol could be used to treat ACM. The original research articles offer limited evidence of atenolol's effectiveness in reducing arrhythmias associated with myocardial infarction, but suggest potential benefits in managing exercise-induced arrhythmias. The clinical case reports, particularly the one involving a family with exercise-induced ventricular arrhythmias, support atenolol's efficacy in reducing such arrhythmias, relevant to ACM. Based on the majority of the evidence from original research and supported by clinical case reports, atenolol shows some promise in managing exercise-induced arrhythmias, which is pertinent for Arrhythmogenic Cardiomyopathy. However, its overall effectiveness and utility specifically for ACM require further validation through clinical trials and wet lab experiments.

**Figure 4. RUGGED Responses.** Question-answer pairs between user queries and RUGGED system responses are shown. In the upper left panel, questions 1-5 retrieve information by constructing queries to the knowledge graph database to formulate evidence-based response. In the upper left panel, the query prompts a prediction analysis which is executed and processed automatically by the system with key findings succinctly summarized. In the bottom panel, literature evidence from the defined text data corpus are retrieved as relevant supporting evidence to verify and support the predicted finding.

| Steps | Description | Time |
|---|---|---|
| *Accessing Biomedical Knowledge* | | *30% total* |
| Prepare biomedical literature corpus | Connect to PubMed and PubMed Central, download and parse publication data for downstream tasks. | 20% |
| Prepare knowledge base data | Connect to biomedical knowledge bases, download and parse necessary information for downstream tasks. | 5% |
| *Information Extraction* | | *30% total* |
| CaseOLAP LIFT Text Mining Analysis | Identify high level disease-protein relationships within the biomedical text corpus. | 25% |
| Knowledge Graph Construction | Connect and integrate disparate information from biomedical knowledge bases into a unified knowledge graph. | 5% |
| *Prediction Analysis* | | *10% total* |
| Train Graph Neural Network | Train the model on the biomedical knowledge graph data to learn hidden patterns within the graph. | 5% |
| Relevance Ranking Analysis | Apply explainability module to highlight the most pertinent nodes and edges relevant to study disease. | 2.5% |
| Link Prediction | Utilize explainability module to identify key nodes and edges contributing to new predicted edges. | 2.5% |
| *Hypothesis Generation and/or Validation* | | *30% total* |
| Database Setup for Retrieval Augmented Generation | Initialize the graph database for querying the knowledge graph and the vector database for text retrieval. | 25% |
| Hypothesis Exploration | Enable user interaction with RUGGED to access and scrutinize relevant information for hypothesis exploration. | 5% |

**Table 1. Workflow and rate limiting steps.** This table provides rough estimates of the computational time required for each stage of the workflow. Rate-limiting steps include accessing, extracting, and indexing biomedical knowledge necessary for retrieval-augmented generation. Hypothesis exploration may be repeated continuously without the need to re-execute rate-limiting steps.

| Disease Category | MeSH Tree Numbers | # PMIDs | # Original Contributions | # Review Articles |
|---|---|---|---|---|
| Cardiomyopathies (CM) | C14.280.238<br>C14.280.434 | 132,531 | 102,337 | 19,942 |
| Cardiac Arrhythmias (ARR) | C14.280.067<br>C23.550.073 | 125,286 | 92,374 | 13,854 |
| Congenital Heart Defects (CHD) | C14.280.400 | 82,006 | 54,023 | 6,379 |
| Heart Valve Diseases (VD) | C14.280.484 | 72,016 | 50,119 | 5,743 |
| Myocardial Ischemia (IHD) | C14.280.647 | 256,986 | 210,042 | 30,223 |
| Cardiac Conduction System Disease (CCD) | C14.280.123 | 53,050 | 35,399 | 4,363 |
| Ventricular Outflow Obstruction (VOO) | C14.280.955 | 22,244 | 15,504 | 1,686 |
| Other Heart Diseases (OTH) | C14.280.195<br>C14.280.282<br>C14.280.383<br>C14.280.470<br>C14.280.945<br>C14.280.459<br>C14.280.720 | 114,085 | 77,302 | 11,799 |
| **Union** | | 635,696 | 478,404 | 69,690 |

**Table 2. Biomedical Literature statistics.** This table details the study disease categories with their corresponding MeSH tree numbers and the number of PubMed documents retrieved from through May 2024, used as input for text mining. A subset of these publications, consisting of original contribution research articles and review articles, is indexed into a vector database for retrieval by RUGGED during hypothesis generation.

| Category | Number of Nodes | Number of Edges | Data Source(s) |
| --- | --- | --- | --- |
| Anatomy | 5,049 | 122,533 | Bgee[40], PubMed, MeSH[41], Uberon[42, 43] |
| Biological Process | 27,047 | 108,106 | Gene Ontology[44] |
| Cellular Component | 4,057 | 52,238 | Gene Ontology |
| Compound | 27,278 | 3,292,028 | DrugBank[45], MeSH, CTD[46], UMLS[47], KEGG[48], TTD[49], SIDER[50], Inxight Drugs[51], Hetionet[52], PathFX[53], MyChem.info[54] |
| Disease | 21,938 | 311,773 | PubMed, MeSH, DisGeNET[55], SIDER, ClinVar[56], ClinGen[57], PharmGKB[58], MyDisease.info[54], PathFX, UMLS, OMIM[59], Mondo[60], DOID[61], KEGG |
| Drug Class | 5,721 | 8,283 | ATC |
| Gene | 29,810 | 943,419 | HGNC[62], GRNdb[63], KEGG, ClinVar, ClinGen, SMPDB[64], DisGeNET, PharmGKB, MyGene.info |
| Molecular Function | 11,151 | 47,086 | Gene Ontology |
| Pathway | 52,012 | 234,944 | Reactome[65], KEGG, SMPDB |
| Protein | 20,740 | 1,074,809 | UniProt[66], Reactome, TTD, SMPDB, STRING[67], HGNC |
| Reaction | 14,647 | 128,038 | Reactome |
| Subtotal | 219,450 | 6,323,257 | |
| Text-mining Associations | 8 | 4,670 | |
| Total | 219,458 | 6,327,927 | |

**Table 3. Knowledge Graph Statistics.** This table details 11 broad biomedical categories comprising the constructed Know2BIO knowledge graph, enriched with additional edges derived from text mining analysis and predictive analysis. The resulting knowledge graph and predictions are managed by the Neo4j graph database for retrieval by RUGGED during hypothesis generation.

|  | Accuracy | Precision | Recall | F1-score | AUROC | AUPRC |
|---|---|---|---|---|---|---|
| Validation | 0.7508 | 0.6732 | 0.9749 | 0.7964 | 0.9634 | 0.9710 |
| Test | 0.7520 | 0.6741 | 0.9755 | 0.7973 | 0.9642 | 0.9716 |

**Table 4. Explainable AI Model Evaluation.** This table reports the evaluation metrics for the knowledge graph link-prediction using a two-layer graph convolutional neural network. Graph edges were partitioned into 85% training, 5% validation, and 10% test datasets. Accuracy indicates the proportion of correctly classified predictions. Precision measures the proportion of correct positive predictions out of all positive predictions, while recall measures the proportion of correct positive predictions out of all positive edges. The F1-score is the harmonic mean of precision and recall. AUROC describes how well the model distinguishes between positive and negative predictions. AUPRC measures the trade-off between precision and recall at varying thresholds, with higher values indicating better performance.

**DISCUSSION:**

The RUGGED protocol leverages modern language models with up-to-date knowledge to empower investigators to dynamically explore the evolving biomedical landscape and uncover new knowledge. This protocol should be executed in the outlined sequence. **Sections 2 and 3** are essential for preparing biomedical knowledge, while **Section 5** prepares the data for retrieval-augmented generation and user interaction with the LLM system. Some time-intensive steps may run concurrently. For example, creating the Neo4j graph (**Section 5a**) can begin during prediction analysis (**Section 4**), and indexing can begin after constructing the knowledge graph (**Section 2b**) and text-mining (**Section 3e**). These steps must be re-run to integrate intermediate results. While designed for biomedical information retrieval, this protocol can handle any text and graph data, such as in-house data, clinical notes, or electronic health records. Data formatting details are in **Section 4**.

Our platform relies on several interconnected technologies, including language models, graph databases, and vector databases. Test scripts are included in the *test* folder in the GitHub repository and are useful to verify these services are running properly. External services such as OpenAI's LLM and Neo4j AuraDB incur user costs, with prices subject to change by the vendor. These optional services have no-cost alternatives using locally hosted services, requiring only sufficient computational resources. However, these alternatives may reduce model performance with improved convenience, making these alternatives unsuitable for some use cases.

With the rapidly evolving LLM landscape, new landmark models and task-specific models are released regularly. At the time of this publication, the most appropriate models were chosen for the task. Users can choose which LLM to use and update the configuration file accordingly by visiting the Hugging Face repository (see **Section 1.iv.2.a**) for open sourced models, or by visiting the LLM service provider website of choice (e.g., OpenAI). These models can be selected depending on their relevance to a particular use case. For example, incorporating models focused on ensuring model responses are fair, censored, and free of hate speech, into this workflow is essential for public-facing models. Furthermore, prompt engineering is necessary to achieve optimal behavior from the LLM. While prompt engineering is largely handled within the RUGGED workflow, effective prompts for one LLM may be suboptimal for another, particularly if the models are fine-tuned on different materials. Users can edit prompts used within the RUGGED workflow in the *configuration* folder, within the *prompts.json* file.

While RAG systems aim to reduce hallucinations in LLMs by grounding responses in evidence, these models can still produce incorrect information. This often occurs when retrieved information exceeds the model's context window, causing it to 'forget' some details. Choosing a suitable LLM model helps to mitigate this issue. For instance, GPT-4o has a context limit of 128k tokens, significantly more than GPT-3.5 Turbo's 16k token limit, albeit at a higher cost. Furthermore, LLMs fine-tuned with specific domain knowledge can potentially enhance accuracy in biomedical applications. Despite these measures, it is essential to cross-check the information before proceeding with costly wet lab experiments.

RUGGED leverages explainable AI within a Retrieval-Augmented Generation (RAG) pipeline to make link predictions, identifying both reliable and previously undiscovered relationships. While traditional RAG systems rely on bulk similarity-based retrieval, our approach connects explainability with a targeted response augmentation. **Table 4** highlights the model's strong performance, demonstrating high recall (validation: 0.9748, test: 0.9755) and balanced F1-scores (validation: 0.7964, test: 0.7973), indicating reliability in identifying true positives, albeit with a higher rate of false positives. The model's robustness is further supported by its AUROC (validation: 0.9634, test: 0.9642) and AUPRC (validation: 0.9710, test: 0.9716) values. Precision (validation: 0.6732, test: 0.6741), however, could benefit from threshold tuning, including detailed node features, or improved handling of class imbalance. The model's effectiveness is highly dependent on the input knowledge graph; overfitting is a risk with smaller graphs, while larger graphs demand greater computational resources.

RUGGED's primary use is hypothesis generation to investigate various hidden relationships, such as disease mechanisms, drug treatments, and speeding up literature triage. To reduce computational burden, many applications can be hosted on a server (e.g., AWS or computational server) and configured to update periodically with the latest information. This workflow can also be adapted to serve specific applications such as serving as a platform to include patient data with local models to uphold security, privacy, and confidentiality. The system can be configured to assist with e-Learning to explain concepts or knowledge at different levels, such as explaining research findings or knowledge base information to the public or students.

**ACKNOWLEDGMENTS:**
This work was supported in part by NIH 1U54HG012517-01 to P.P., NIH T32 EB016640 to A.R.P., NIH T32 HL13945 to A.R.P. and D.S., National Science Foundation Research Traineeship (NRT) 1829071 to A.R.P. and D.S., and the TC Laubisch Endowment to P.P. at UCLA.

**DISCLOSURES:**
The authors have nothing to disclose.

## Materials

| Name | Company | Catalog Number | Comments |
| --- | --- | --- | --- |
| Software - Docker | Docker | N/A | docker.com |
| Software - Neo4j | Neo4j | N/A | neo4j.com |
| Software - OpenAI (optional) | OpenAI | N/A | openai.com |

# Supplementary Materials

## User: What are the currently prescribed drugs classified as beta blockers, antiarrhythmic drugs, and antifibrotic drugs?

## RUGGED:

Based on the classifications of drugs you provided, our system identified relevant ATC codes for beta blockers, antiarrhythmic drugs, and antifibrotic drugs, which are respectively ATC_Class:C07, ATC_Class:C01B, and ATC_Class:L01X.

Using these classifications, a Cypher query was constructed to extract a list of drugs for each class from a knowledge graph:
- **Beta Blockers (36 drugs):** Included compounds such as DB00187, DB00195, DB00264, and so forth, up to DB13775.
- **Antiarrhythmic Drugs (29 drugs):** Included DB00204, DB00280, DB00281, and additional compounds up to DB15300.
- **Antifibrotic Drugs (53 drugs):** Started with DB00317 and included others such as DB00398, DB00530, through to DB14723.

This approach systematically identified and listed the currently prescribed drugs under each specified category by employing the structured query capabilities of a knowledge graph.

cypher command used to access this information:

```
MATCH (d:DrugBank_Compound)-[:`-compound_classified_as_drug_class->`]->(a:ATC_Class)
WHERE a.name IN ['ATC_Class:C07', 'ATC_Class:C01B', 'ATC_Class:L01X']
RETURN a.name AS ATC_Code, COLLECT(DISTINCT d.name) AS Drugs
ORDER BY ATC_Code
```

# Question 2:

## User: Which drug is the most commonly prescribed in each category?

## RUGGED:

Using a Cypher query, we analyzed the most connected drug compounds in three categories—Beta Blockers, Antiarrhythmics, and Antifibrotics—from a knowledge graph to identify their frequency and relevance in medical treatments. The drugs were ranked by their connections, highlighting their widespread prescription.

**Most Commonly Prescribed Drugs by Category:**

1. **Beta Blockers**
    - **Drug Name:** Pindolol
    - **Connections:** 16,668
    - **Description:** Pindolol is primarily used in the treatment of hypertension, arrhythmia, and related cardiovascular conditions. It functions as a non-selective beta-blocker, impacting both beta-1 and beta-2 adrenergic receptors. It was first approved by the FDA in 1982, reflecting its longstanding use in managing heart-related ailments.
2. **Antiarrhythmics**
    - **Drug Name:** Quinidine
    - **Connections:** 20,216
    - **Description:** Quinidine serves as a significant agent in treating arrhythmias such as atrial fibrillation and flutter, and ventricular arrhythmias. It is known for its capacity to restore normal sinus rhythm and for prolonging the cellular action potential by blocking sodium and potassium currents.
3. **Antifibrotics**
    - **Drug Name:** Sorafenib
    - **Connections:** 15,888
    - **Description:** Sorafenib is a kinase inhibitor used in the treatment of unresectable liver carcinoma, advanced renal carcinoma, and differentiated thyroid carcinoma. Its approval in 2007 marked it as a critical drug for targeting cell proliferation and angiogenesis in cancer therapy.

cypher command used to access this information:

```
// Top Beta Blocker
MATCH (d:DrugBank_Compound)-[r]-()
WHERE any(key in [
    'DB00187', 'DB00195', 'DB00264', 'DB00335', 'DB00373', 'DB00381', 'DB00489',
    'DB00521', 'DB00571', 'DB00598', 'DB00612', 'DB00866', 'DB00945', 'DB00960',
    'DB01023', 'DB01115', 'DB01136', 'DB01193', 'DB01203', 'DB01214', 'DB01295',
    'DB01297', 'DB01359', 'DB01580', 'DB04846', 'DB04861', 'DB08807', 'DB08808',
    'DB09083', 'DB11770', 'DB12212', 'DB13443', 'DB13508', 'DB13530', 'DB13757', 'DB13775'
    ] WHERE d.name CONTAINS key)
RETURN d.name AS DrugName, COUNT(r) AS Connections, 'Beta Blockers' AS Category
ORDER BY Connections DESC LIMIT 1
UNION ALL
// Top Antiarrhythmic
MATCH (d:DrugBank_Compound)-[r]-()
WHERE any(key in [
    'DB00204', 'DB00280', 'DB00281', 'DB00308', 'DB00379', 'DB00680', 'DB00908',
    'DB01035', 'DB01056', 'DB01118', 'DB01158', 'DB01182', 'DB01195', 'DB01228',
    'DB01426', 'DB01429', 'DB04855', 'DB06200', 'DB06217', 'DB06727', 'DB13358',
    'DB13555', 'DB13645', 'DB13651', 'DB13652', 'DB13653', 'DB15300'
    ] WHERE d.name CONTAINS key)
RETURN d.name AS DrugName, COUNT(r) AS Connections, 'Antiarrhythmics' AS Category
ORDER BY Connections DESC LIMIT 1
UNION ALL
// Top Antifibrotic
MATCH (d:DrugBank_Compound)-[r]-()
WHERE any(key in [
    'DB00317', 'DB00398', 'DB00530', 'DB00619', 'DB01254', 'DB01259', 'DB01268',
```

```
        'DB01590', 'DB04849', 'DB04868', 'DB05239', 'DB05294', 'DB06233', 'DB06287',
        'DB06589', 'DB06595', 'DB06616', 'DB06626', 'DB08865', 'DB08875', 'DB08877',
        'DB08881', 'DB08896', 'DB08901', 'DB08911', 'DB08912', 'DB08916', 'DB09053',
        'DB09063', 'DB09073', 'DB09078', 'DB09079', 'DB09330', 'DB11363', 'DB11526',
        'DB11703', 'DB11718', 'DB11730', 'DB11737', 'DB11800', 'DB11828', 'DB11907',
        'DB11963', 'DB11967', 'DB11986', 'DB12001', 'DB12130', 'DB12141', 'DB12267',
        'DB12500', 'DB12874', 'DB13164', 'DB14723'
        ] WHERE d.name CONTAINS key)
RETURN d.name AS DrugName, COUNT(r) AS Connections, 'Antifibrotics' AS Category
ORDER BY Connections DESC LIMIT 1;
```

# Question 3:

### User: Which of these drugs are relevant for the treatment of Arrhythmogenic Cardiomyopathy (ACM) and/or Dilated Cardiomyopathy (DCM)?

## RUGGED:

In response to the inquiry regarding the relevance of specific drugs for treating Arrhythmogenic Cardiomyopathy (ACM) and Dilated Cardiomyopathy (DCM), our system identified relevant MeSH disease matches for ACM and DCM as MeSH_Disease:D002312 and MeSH_Disease:D002311, respectively. We constructed a Cypher query to examine a list of drugs within the categories of Beta Blockers, Antiarrhythmics, and Antifibrotics. The query confirmed the presence of several key drugs for both diseases, providing a list of compounds active in the treatment of these cardiac conditions. Below is a brief description of each of the 7 resulting drug identified in the analysis:

**Beta Blockers**

1. **Metoprolol (DB00264)** is known to treat both ACM and DCM.

- **Description:** Metoprolol is a selective beta-1 blocker widely used for hypertension, angina, and to reduce mortality following myocardial infarction. Available in immediate and extended-release forms, it's preferred for its cardiovascular benefits and is extensively prescribed in various countries.
2. **Propranolol (DB00571)** is known to treat ACM.
    - **Description:** A non-selective beta adrenergic antagonist, Propranolol is used for a variety of conditions including hypertension, angina, and arrhythmias. It's known for its high binding affinity to beta adrenergic receptors and was one of the first of its kind to be approved by the FDA.
    - Known to treat
3. **Nifedipine (DB01115)** is known to treat both ACM and DCM.
    - **Description:** This dihydropyridine calcium channel blocker is primarily used for angina and hypertension management. Nifedipine is noted for its role in cardiovascular disease management with its approval dating back to 1981.
4. **Carvedilol (DB01136)** is known to treat both ACM and DCM.
    - **Description:** Carvedilol acts as a non-selective beta and alpha-1 adrenoceptor blocker, making it effective for treating heart failure, hypertension, and left ventricular dysfunction post-myocardial infarction. Its dual action offers an advantage in combination therapies.
5. **Atenolol (DB00335)** is known to treat DCM.
    - Description: Atenolol is a beta-1 selective blocker used for treating hypertension, chronic angina, and reducing mortality post-myocardial infarction. Despite its popularity, atenolol may only modestly reduce cardiovascular risk and mortality in hypertensive patients.

**Antiarrhythmics**

1. **Disopyramide (DB00280)** is known to treat ACM.
    - **Description:** Employed in the treatment of life-threatening ventricular arrhythmias, Disopyramide is a class 1A antiarrhythmic that stabilizes the cardiac membrane. It is used in severe cases where potent action is required.
2. **Amiodarone (DB01118)** is known to treat both ACM and DCM.
    - **Description:** A class III antiarrhythmic, Amiodarone is indicated for recurrent ventricular fibrillation and unstable ventricular tachycardia. Known for its effectiveness in managing severe arrhythmias, it also carries a risk of serious side effects, limiting its use to critical cases.

No Antifibrotics were identified within the knowledge graph.

This summarized analysis, supported by a Cypher query, confirms the involvement of these drugs in the treatment protocols for ACM and DCM, highlighting their critical roles in managing these complex conditions.

cypher command used to access this information:

```
// Define the list of compounds to check
WITH [
    'DrugBank_Compound:DB00187',
'DrugBank_Compound:DB00195', 'DrugBank_Compound:DB00264',
'DrugBank_Compound:DB00335', 'DrugBank_Compound:DB00373',
'DrugBank_Compound:DB00381', 'DrugBank_Compound:DB00489',
'DrugBank_Compound:DB00521', 'DrugBank_Compound:DB00571',
'DrugBank_Compound:DB00598', 'DrugBank_Compound:DB00612',
'DrugBank_Compound:DB00866', 'DrugBank_Compound:DB00945',
'DrugBank_Compound:DB00960', 'DrugBank_Compound:DB01023',
'DrugBank_Compound:DB01115', 'DrugBank_Compound:DB01136',
'DrugBank_Compound:DB01193', 'DrugBank_Compound:DB01203',
'DrugBank_Compound:DB01214', 'DrugBank_Compound:DB01295',
'DrugBank_Compound:DB01297', 'DrugBank_Compound:DB01359',
'DrugBank_Compound:DB01580', 'DrugBank_Compound:DB04846',
'DrugBank_Compound:DB04861', 'DrugBank_Compound:DB08807',
'DrugBank_Compound:DB08808', 'DrugBank_Compound:DB09083',
'DrugBank_Compound:DB11770', 'DrugBank_Compound:DB12212',
'DrugBank_Compound:DB13443', 'DrugBank_Compound:DB13508',
'DrugBank_Compound:DB13530', 'DrugBank_Compound:DB13757',
'DrugBank_Compound:DB13775', // Beta Blockers
    'DrugBank_Compound:DB00204',
'DrugBank_Compound:DB00280', 'DrugBank_Compound:DB00281',
'DrugBank_Compound:DB00308', 'DrugBank_Compound:DB00379',
'DrugBank_Compound:DB00680', 'DrugBank_Compound:DB00908',
'DrugBank_Compound:DB01035', 'DrugBank_Compound:DB01056',
'DrugBank_Compound:DB01118', 'DrugBank_Compound:DB01158',
'DrugBank_Compound:DB01182', 'DrugBank_Compound:DB01195',
'DrugBank_Compound:DB01228', 'DrugBank_Compound:DB01426',
'DrugBank_Compound:DB01429', 'DrugBank_Compound:DB04855',
'DrugBank_Compound:DB06200', 'DrugBank_Compound:DB06217',
'DrugBank_Compound:DB06727', 'DrugBank_Compound:DB13358',
'DrugBank_Compound:DB13555', 'DrugBank_Compound:DB13645',
'DrugBank_Compound:DB13651', 'DrugBank_Compound:DB13652',
```

```
    'DrugBank_Compound:DB13653', 'DrugBank_Compound:DB15300',
// Anti-arrythmic
    'DrugBank_Compound:DB00317',
    'DrugBank_Compound:DB00398', 'DrugBank_Compound:DB00530',
    'DrugBank_Compound:DB00619', 'DrugBank_Compound:DB01254',
    'DrugBank_Compound:DB01259', 'DrugBank_Compound:DB01268',
    'DrugBank_Compound:DB01590', 'DrugBank_Compound:DB04849',
    'DrugBank_Compound:DB04868', 'DrugBank_Compound:DB05239',
    'DrugBank_Compound:DB05294', 'DrugBank_Compound:DB06233',
    'DrugBank_Compound:DB06287', 'DrugBank_Compound:DB06589',
    'DrugBank_Compound:DB06595', 'DrugBank_Compound:DB06616',
    'DrugBank_Compound:DB06626', 'DrugBank_Compound:DB08865',
    'DrugBank_Compound:DB08875', 'DrugBank_Compound:DB08877',
    'DrugBank_Compound:DB08881', 'DrugBank_Compound:DB08896',
    'DrugBank_Compound:DB08901', 'DrugBank_Compound:DB08911',
    'DrugBank_Compound:DB08912', 'DrugBank_Compound:DB08916',
    'DrugBank_Compound:DB09053', 'DrugBank_Compound:DB09063',
    'DrugBank_Compound:DB09073', 'DrugBank_Compound:DB09078',
    'DrugBank_Compound:DB09079', 'DrugBank_Compound:DB09330',
    'DrugBank_Compound:DB11363', 'DrugBank_Compound:DB11526',
    'DrugBank_Compound:DB11703', 'DrugBank_Compound:DB11718',
    'DrugBank_Compound:DB11730', 'DrugBank_Compound:DB11737',
    'DrugBank_Compound:DB11800', 'DrugBank_Compound:DB11828',
    'DrugBank_Compound:DB11907', 'DrugBank_Compound:DB11963',
    'DrugBank_Compound:DB11967', 'DrugBank_Compound:DB11986',
    'DrugBank_Compound:DB12001', 'DrugBank_Compound:DB12130',
    'DrugBank_Compound:DB12141', 'DrugBank_Compound:DB12267',
    'DrugBank_Compound:DB12500', 'DrugBank_Compound:DB12874',
    'DrugBank_Compound:DB13164', 'DrugBank_Compound:DB14723'
//Anti-fibrotics
] AS drugList

// Match diseases from a specific list
MATCH (disease:MeSH_Disease)
WHERE disease.name IN [
    'MeSH_Disease:D002312','MeSH_Disease:D002311'
]

// Match the drugs that treat these diseases
OPTIONAL MATCH (m)-[:`-treats->`]->(disease)
```

```
WHERE m.name IN drugList

// Return results
RETURN disease.name AS Disease, COLLECT(DISTINCT m.name) AS
Compounds
```

# Question 4:

### User: What are the molecular targets of these drugs?

## RUGGED:

Based on the results of your query, I can provide information about the molecular targets of the specified drugs, which are Beta Blockers and Antiarrhythmics. Here are the details:

**Beta Blockers:**
1. DB00264 (Metoprolol) has 2 molecular targets:
   - UniProt:P08588 (Beta-1 adrenergic receptor)
   - UniProt:P07550 (Beta-2 adrenergic receptor)

2. DB00571 (Propranolol) has 5 molecular targets:
   - UniProt:P08588 (Beta-1 adrenergic receptor)
   - UniProt:P28222 (5-hydroxytryptamine receptor 1B)
   - UniProt:P08908 (5-hydroxytryptamine receptor 1A)
   - UniProt:P13945 (Beta-3 adrenergic receptor)
   - UniProt:P07550 (Beta-2 adrenergic receptor)

3. DB01115 (Nifedipine) has 8 molecular targets:
   - UniProt:Q13936 (Voltage-dependent T-type calcium channel subunit alpha-1C)
   - UniProt:O43497 (Voltage-dependent T-type calcium channel subunit alpha-1G)
   - UniProt:Q9UK17 (Potassium voltage-gated channel subfamily D member 3)
   - UniProt:Q08289 (Voltage-dependent L-type calcium channel subunit beta-2)
   - UniProt:Q13698 (Voltage-dependent L-type calcium channel subunit alpha-1S)
   - UniProt:Q01668 (Voltage-dependent L-type calcium channel subunit alpha-1D)
   - UniProt:P0DP23 (Calmodulin-1)
   - UniProt:O75469 (Nuclear receptor subfamily 1 group I member 2)

4. DB01136 (Carvedilol) has 18 molecular targets:
    - UniProt:P08588 (Beta-1 adrenergic receptor)
    - UniProt:P63252 (Inward rectifier potassium channel 2)
    - UniProt:P48050 (Inward rectifier potassium channel 4)
    - UniProt:Q16665 (Hypoxia-inducible factor 1-alpha)
    - UniProt:P16581 (E-selectin)
    - UniProt:P08913 (Alpha-2A adrenergic receptor)
    - UniProt:P18089 (Alpha-2B adrenergic receptor)
    - UniProt:P18825 (Alpha-2C adrenergic receptor)
    - UniProt:P35368 (Alpha-1B adrenergic receptor)
    - UniProt:P25100 (Alpha-1D adrenergic receptor)
    - UniProt:P19320 (Vascular cell adhesion protein 1)
    - UniProt:Q12809 (Potassium voltage-gated channel subfamily H member 2)
    - UniProt:P17302 (Gap junction alpha-1 protein)
    - UniProt:P16860 (Natriuretic peptides B)
    - UniProt:P15692 (Vascular endothelial growth factor A, long form)
    - UniProt:P07550 (Beta-2 adrenergic receptor)
    - UniProt:O95298 (NADH dehydrogenase [ubiquinone] 1 subunit C2)
    - UniProt:P35349 (Metabotropic glutamate receptor 6)

5. DB00335 (Atenolol) has 2 molecular targets:
    - UniProt:Q15822 (Neuronal acetylcholine receptor subunit alpha-2)
    - UniProt:P08172 (Muscarinic acetylcholine receptor M2)

**Antiarrhythmics:**
1. DB00280 (Disopyramide) has 8 molecular targets:
    - UniProt:Q14524 (Sodium channel protein type 5 subunit alpha)
    - UniProt:P19652 (Alpha-1-acid glycoprotein 2)
    - UniProt:Q12809 (Potassium voltage-gated channel subfamily H member 2)
    - UniProt:Q9UK17 (Potassium voltage-gated channel subfamily D member 3)
    - UniProt:Q9NZV8 (Potassium voltage-gated channel subfamily D member 2)
    - UniProt:P20309 (Muscarinic acetylcholine receptor M3)
    - UniProt:P08172 (Muscarinic acetylcholine receptor M2)
    - UniProt:P11229 (Muscarinic acetylcholine receptor M1)

2. DB01118 (Amiodarone) has 7 molecular targets:
    - UniProt:Q12809 (Potassium voltage-gated channel subfamily H member 2)
    - UniProt:Q86YN6 (Peroxisome proliferator-activated receptor gamma coactivator 1-beta)
    - UniProt:Q07869 (Peroxisome proliferator-activated receptor alpha)

- UniProt:P37231 (Peroxisome proliferator-activated receptor gamma)
- UniProt:Q13936 (Voltage-dependent L-type calcium channel subunit alpha-1C)
- UniProt:Q9P0X4 (Voltage-dependent T-type calcium channel subunit alpha-1I)
- UniProt:P08588 (Beta-1 adrenergic receptor)

The molecular targets of the specified drugs primarily include various adrenergic receptors, calcium channels, and potassium channels. Beta blockers like Metoprolol, Propranolol, and Carvedilol target multiple adrenergic receptors, contributing to their effectiveness in managing hypertension and cardiovascular conditions. Antiarrhythmics like Disopyramide and Amiodarone target a diverse range of ion channels and receptors, reflecting their role in managing arrhythmias. Carvedilol stands out with the highest number of molecular targets, indicating its broad pharmacological activity. The overall trend shows that these drugs interact with a combination of adrenergic receptors, calcium channels, and potassium channels, which are crucial in cardiovascular and nervous system functions.

cypher command used to access this information:

```
MATCH (drug:DrugBank_Compound)-[:`-drug_targets_protein->`]
->(target:UniProt)
WHERE drug.name IN ['DrugBank_Compound:DB00264',
'DrugBank_Compound:DB00571', 'DrugBank_Compound:DB01115',
'DrugBank_Compound:DB01136','DrugBank_Compound:DB01135',
'DrugBank_Compound:DB00280', 'DrugBank_Compound:DB01118']
RETURN drug.name AS Drug, COLLECT(target.name) AS Targets
```

# Question 5:

### User: How do the molecular targets of these drugs interact with each other?

### RUGGED:

Based on the results of your query, here's a summary of how the molecular targets of the specified drugs interact with each other:

1. UniProt:P08588 (Beta-1 adrenergic receptor) interacts with 24 proteins, including Beta-2 adrenergic receptor and several other signaling molecules and receptors.

2. UniProt:P28222 (5-hydroxytryptamine receptor 1B) interacts with 12 proteins, which include other receptors and signaling proteins.
3. UniProt:P08908 (5-hydroxytryptamine receptor 1A) interacts with 20 proteins, encompassing various receptors and signaling molecules.
4. UniProt:P13945 (Beta-3 adrenergic receptor) interacts with 3 proteins, including potassium channels.
5. UniProt:P07550 (Beta-2 adrenergic receptor) interacts with 37 proteins, indicating extensive interactions with various receptors and signaling molecules.
6. UniProt:Q13936 (Voltage-dependent T-type calcium channel subunit alpha-1C) interacts with 78 proteins, showing a wide range of interactions including other calcium channels and signaling molecules.
7. UniProt:O43497 (Voltage-dependent T-type calcium channel subunit alpha-1G) interacts with 11 proteins, involving various channels and receptors.
8. UniProt:Q9UK17 (Potassium voltage-gated channel subfamily D member 3) interacts with 21 proteins, including various ion channels and signaling proteins.
9. UniProt:Q08289 (Voltage-dependent L-type calcium channel subunit beta-2) interacts with 45 proteins, reflecting its extensive role in cellular signaling and interactions with other channels.
10. UniProt:Q13698 (Voltage-dependent L-type calcium channel subunit alpha-1S) interacts with 12 proteins, including other calcium channels and related signaling molecules.
11. UniProt:Q01668 (Voltage-dependent L-type calcium channel subunit alpha-1D) interacts with 74 proteins, showing extensive interactions within cellular signaling pathways.
12. UniProt:P0DP23 (Calmodulin-1) interacts with 58 proteins, highlighting its critical role in calcium signaling and interactions with various kinases and receptors.
13. UniProt:O75469 (Nuclear receptor subfamily 1 group I member 2) interacts with 19 proteins, indicating its role in nuclear signaling and gene regulation.
14. UniProt:Q12809 (Potassium voltage-gated channel subfamily H member 2) interacts with 22 proteins, involved in various signaling pathways and ion channel regulation.
15. UniProt:Q86YN6 (Peroxisome proliferator-activated receptor gamma coactivator 1-beta) interacts with 3 proteins involved in metabolic regulation.
16. UniProt:Q07869 (Peroxisome proliferator-activated receptor alpha) interacts with 72 proteins, playing a significant role in lipid metabolism and regulation.
17. UniProt:P37231 (Peroxisome proliferator-activated receptor gamma) interacts with 143 proteins, indicating its extensive role in metabolic regulation and gene expression.
18. UniProt:Q9P0X4 (Voltage-dependent T-type calcium channel subunit alpha-1I) interacts with 3 proteins involved in calcium signaling.
19. UniProt:P63252 (Inward rectifier potassium channel 2) interacts with 4 proteins, reflecting its role in maintaining resting membrane potential.
20. UniProt:P48050 (Inward rectifier potassium channel 4) interacts with 4 proteins, involved in regulating membrane potential.
21. UniProt:Q16665 (Hypoxia-inducible factor 1-alpha) interacts with 140 proteins, showing its central role in cellular response to hypoxia.
22. UniProt:P16581 (E-selectin) interacts with 22 proteins, involved in inflammation and immune response.

23. UniProt:P08913 (Alpha-2A adrenergic receptor) interacts with 12 proteins, reflecting its role in neurotransmission and signaling.
24. UniProt:P18089 (Alpha-2B adrenergic receptor) interacts with 2 proteins, indicating specific signaling interactions.
25. UniProt:P35368 (Alpha-1B adrenergic receptor) interacts with 35 proteins, involved in various signaling pathways.
26. UniProt:P25100 (Alpha-1D adrenergic receptor) interacts with 3 proteins, indicating specific receptor interactions.
27. UniProt:P19320 (Vascular cell adhesion protein 1) interacts with 13 proteins, highlighting its role in cell adhesion and signaling.
28. UniProt:P17302 (Gap junction alpha-1 protein) interacts with 60 proteins, showing its extensive role in cell-cell communication.
29. UniProt:P16860 (Natriuretic peptides B) interacts with 6 proteins, involved in cardiovascular regulation.
30. UniProt:P15692 (Vascular endothelial growth factor A, long form) interacts with 67 proteins, highlighting its critical role in angiogenesis and signaling.
31. UniProt:O95298 (NADH dehydrogenase [ubiquinone] 1 subunit C2) interacts with 2 proteins, reflecting its role in mitochondrial function.
32. UniProt:P35348 (Alpha-1A adrenergic receptor) interacts with 8 proteins involved in neurotransmission.
33. UniProt:Q15822 (Neuronal acetylcholine receptor subunit alpha-2) interacts with 2 proteins, indicating specific neuronal signaling interactions.
34. UniProt:P08172 (Muscarinic acetylcholine receptor M2) interacts with 9 proteins involved in cholinergic signaling.
35. UniProt:P19652 (Alpha-1-acid glycoprotein 2) interacts with 11 proteins, reflecting its role in binding and transporting various molecules in the blood.
36. UniProt:Q9NZV8 (Potassium voltage-gated channel subfamily D member 2) interacts with 25 proteins, indicating extensive signaling interactions.
37. UniProt:P20309 (Muscarinic acetylcholine receptor M3) interacts with 2 proteins involved in cholinergic signaling.
38. UniProt:P11229 (Muscarinic acetylcholine receptor M1) interacts with 61 proteins, highlighting its significant role in neurotransmission and signaling.

The molecular targets of the specified drugs exhibit extensive and diverse interactions with other proteins, reflecting their roles in complex signaling pathways and physiological processes. Key targets like Beta-1 adrenergic receptor, Beta-2 adrenergic receptor, and various calcium channels demonstrate broad interaction networks, influencing numerous cellular functions. These interactions underscore the multifaceted mechanisms through which these drugs exert their therapeutic effects, particularly in cardiovascular and nervous system regulation.

cypher command used to access this information:

```
MATCH (drug:DrugBank_Compound)-[:`-drug_targets_protein->`]
->(target1:UniProt)
```

```
WHERE drug.name IN ['DrugBank_Compound:DB00264',
'DrugBank_Compound:DB00571', 'DrugBank_Compound:DB01115',
'DrugBank_Compound:DB01136','DrugBank_Compound:DB01135',
'DrugBank_Compound:DB00280', 'DrugBank_Compound:DB01118']
WITH DISTINCT target1
MATCH
(target1)-[r:interacts_with|`-ppi-`]->(target2:UniProt)
RETURN target1.name AS Target1, COLLECT(DISTINCT
target2.name) AS InteractingProteins
```

# Question 6:

### User: What biological functions and cellular processes are associated with the molecular targets of these drugs?

## RUGGED:

This analysis examines the molecular targets and biological effects of seven drugs, including five beta blockers (Metoprolol, Propranolol, Nifedipine, Carvedilol, Atenolol) and two antiarrhythmics (Disopyramide, Amiodarone). By analyzing the Gene Ontology (GO) terms associated with the proteins targeted by these drugs, we can elucidate their high-level biological functions and cellular processes. Understanding these interactions is crucial for comprehending how these drugs exert their therapeutic effects. Based on the results of your query, I can provide detailed information about the biological functions and cellular processes associated with the molecular targets of the specified drugs.

**Beta Blockers**

**DB00264 (Metoprolol)**
Metoprolol targets the Beta-1 adrenergic receptor (UniProt:P08588) and the Beta-2 adrenergic receptor (UniProt:P07550). This drug primarily affects heart rate and vascular smooth muscle contraction, leading to decreased heart rate (GO:0001997) and vasodilation (GO:0042310). By blocking these receptors, metoprolol reduces the workload on the heart and helps to manage hypertension and arrhythmias.

**DB00571 (Propranolol)**
Propranolol targets multiple receptors, including the Beta-1 adrenergic receptor (UniProt:P08588), Beta-2 adrenergic receptor (UniProt:P07550), Beta-3 adrenergic receptor (UniProt:P13945), and two serotonin receptors (UniProt:P28222 and UniProt:P08908). This broad action profile means it not only reduces heart rate and blood pressure but also has effects on the central nervous system, such as anxiety reduction (GO:0051967) and regulation of neurotransmitter secretion (GO:0050795).

**DB01115 (Nifedipine)**
Nifedipine targets various voltage-dependent calcium channels, including those in the L-type and T-type (e.g., UniProt:Q13936, UniProt:O43497). These targets are involved in regulating calcium influx into cardiac and smooth muscle cells, affecting processes like muscle contraction and vascular tone (GO:0007204, GO:0007218). Nifedipine primarily serves to lower blood pressure and relieve angina by causing vasodilation (GO:0042310).

**DB01136 (Carvedilol)**
Carvedilol interacts with a wide range of molecular targets, including adrenergic receptors (Beta-1, Beta-2, Alpha-1), potassium channels, and proteins involved in cell adhesion and vascular function. This broad interaction spectrum allows carvedilol to reduce heart rate and blood pressure (GO:0001997), protect against vascular inflammation (GO:0002523), and improve cardiac output by affecting multiple signaling pathways and reducing the workload on the heart.

**DB00335 (Atenolol)**
Atenolol targets the Beta-1 adrenergic receptor (UniProt:P08588), reducing heart rate and myocardial contractility (GO:0001997, GO:0050873). It also affects neuronal acetylcholine receptor subunit alpha-2 (UniProt:Q15822), influencing neural signaling and contributing to its anti-hypertensive effects (GO:0019228).

**Antiarrhythmics**

**DB00280 (Disopyramide)**
Disopyramide targets various ion channels, including sodium and potassium channels (e.g., UniProt:Q14524, UniProt:Q12809), and muscarinic acetylcholine receptors (UniProt:P20309, UniProt:P08172). By blocking these channels, Disopyramide stabilizes cardiac membranes and prevents abnormal electrical activity, making it effective in treating arrhythmias (GO:0007268, GO:0055117).

**DB01118 (Amiodarone)**
Amiodarone targets multiple ion channels and receptors, including the Beta-1 adrenergic receptor (UniProt:P08588) and several potassium and calcium channels (e.g., UniProt:Q12809, UniProt:Q13936). This broad range of targets allows it to modulate cardiac rhythm by inhibiting abnormal electrical activity and stabilizing cardiac cell membranes (GO:0007268, GO:0051260). Additionally, it influences peroxisome proliferator-activated receptors (UniProt:Q07869, UniProt:P37231), affecting lipid metabolism and energy homeostasis (GO:0019228, GO:0045762).

**Summary**

Beta blockers and antiarrhythmics target various proteins to exert their effects on the cardiovascular system and beyond. Beta blockers like Metoprolol, Propranolol, and Carvedilol primarily interact with adrenergic receptors, leading to reduced heart rate and blood pressure. Additionally, Propranolol and Carvedilol have broader target profiles affecting neurotransmitter regulation and vascular inflammation. Nifedipine, while often classified as a calcium channel blocker, targets multiple calcium channels to reduce vascular tone and alleviate hypertension.

Antiarrhythmics like Disopyramide and Amiodarone stabilize cardiac membranes by targeting ion channels and receptors involved in cardiac excitability. Disopyramide's effects on sodium and potassium channels prevent abnormal electrical activity in the heart, making it effective in treating arrhythmias. Amiodarone's broader target range, including potassium, calcium channels, and adrenergic receptors, provides a multifaceted approach to modulating cardiac rhythm and stabilizing heart function.

cypher command used to access this information:

```
MATCH
(drug:DrugBank_Compound)-[:`-drug_targets_protein->`]->(target:UniProt)
WHERE drug.name IN ['DrugBank_Compound:DB00264',
'DrugBank_Compound:DB00571', 'DrugBank_Compound:DB01115',
'DrugBank_Compound:DB01136','DrugBank_Compound:DB01135',
'DrugBank_Compound:DB00280', 'DrugBank_Compound:DB01118']
WITH DISTINCT target

// Find associated biological functions
OPTIONAL MATCH
(target)-[:enables]->(func:molecular_function)
WITH target, collect(DISTINCT func.name) AS Functions

// Find associated cellular components
OPTIONAL MATCH
(target)-[:located_in]->(comp:cellular_component)
WITH target, Functions, collect(DISTINCT comp.name) AS
CellularComponents

// Find associated biological processes
OPTIONAL MATCH
(target)-[:involved_in]->(proc:biological_process)
```

```
RETURN target.name AS Target, Functions,
CellularComponents, collect(DISTINCT proc.name) AS
BiologicalProcesses
```

## Question 7:

**User: Which of the prescribed beta blockers, antiarrhythmic drugs, and antifibrotic drugs could potentially be used to treat ACM and/or DCM that is not currently known? Please identify the top 3 predictions.**

## RUGGED:

The list of prescribed beta blockers and antiarrhythmic drugs for their known treatments of Arrhythmogenic Cardiomyopathy (ACM) and Dilated Cardiomyopathy (DCM) are explored below. The explainability analysis identified three top predictions for drugs that could potentially treat ACM and/or DCM, which are not currently known for these uses. The top three predictions are:

1. **Propranolol (DB00571) and Dilated Cardiomyopathy (DCM)**: Predicted probability of 0.9828.
2. **Lidocaine (DB00280) and Dilated Cardiomyopathy (DCM)**: Predicted probability of 0.9458.
3. **Atenolol (DB00335) and Arrhythmogenic Cardiomyopathy (ACM)**: Predicted probability of 0.9828.

Below is an example of one such prediction between Atenolol (DB00335) and Arrhythmogenic Cardiomyopathy (ACM).

To address the question of a common protein drug target for Atenolol (DB00335) and Arrhythmogenic Cardiomyopathy (ACM), a link prediction analysis using graph explainability techniques was performed. This involved evaluating the top-scoring relationship between Atenolol (DB00335) and Arrhythmogenic Cardiomyopathy (MeSH_Disease:D002313) by analyzing the most influential edges and paths. The analysis focused on direct and indirect connections, examining their biological implications and assessing the strength and reliability of the predictions. The process involved identifying relevant nodes, analyzing their interactions, and evaluating the potential for these predictions to be validated in wet lab experiments.

**Influential Nodes and Paths**

- (MeSH_Disease:C566005, MeSH_Disease:C566171): With a score of 0.830, this edge suggests a strong association between these disease nodes, implying potential shared pathways or mechanisms relevant to ACM.
- (MeSH_Tree_Disease:C14.280.238, MeSH_Tree_Disease:C14.280.238.160): The score of 0.808 indicates an important relationship within the disease hierarchy, pointing to specific subtypes or related conditions that might share pathophysiological mechanisms with ACM.
- (MeSH_Disease:C566168, MeSH_Disease:C567654): This edge, with a score of 0.627, highlights a connection between diseases that may share common etiological factors with ACM.
- (MeSH_Disease:C567505, MeSH_Disease:C567654): The score of 0.565 for this edge underscores another significant disease relationship, reinforcing the interconnectedness of these conditions.
- (MeSH_Disease:C567505, MeSH_Disease:C566168): This edge (0.333) suggests a moderate association, indicating potential pathways or mechanisms that could be explored for shared disease processes.
- (MeSH_Disease:C563866, MeSH_Disease:C566005): The score of 0.247 points to an additional disease connection, suggesting that these diseases might influence each other or share similar biological pathways.
- (MeSH_Disease:C566171, MeSH_Disease:D002313): With a score of 0.243, this edge highlights a direct relationship with ACM, indicating that diseases in this network might provide insights into ACM pathogenesis.
- (MeSH_Tree_Disease:C14.280.238.160, MeSH_Disease:D002313): This edge (0.217) emphasizes the importance of hierarchical disease relationships in understanding ACM.
- (MeSH_Disease:C566005, MeSH_Disease:D002313): The score of 0.205 suggests a relevant disease-disease interaction that could contribute to understanding ACM mechanisms.
- (ATC_Class:C07BB03, DrugBank_Compound:DB00335): The score of 0.194 highlights the classification and pharmacological context of Atenolol, linking it to its therapeutic use in cardiovascular diseases.

**Potential Biological Implications**
- Disease Mechanisms: The high-scoring edges between Atenolol and ACM, and other diseases, suggest shared pathogenic pathways. Existing edges between ACM and other diseases include the edge "-diseases_share_genes-" between disease nodes, indicating shared biological mechanisms. Exploring these shared gene associations within the knowledge graph can provide deeper insights into the common pathogenic pathways. This could imply that targeting these shared pathways might have therapeutic effects on ACM due to the underlying common mechanisms.
- ATC Classification: The Anatomical Therapeutic Chemical (ATC) classification system categorizes drugs based on their therapeutic use and chemical characteristics. Atenolol is classified under C07BB03, indicating it is a beta-blocker used in cardiovascular diseases. This classification underscores Atenolol's role in modulating cardiovascular functions, aligning with its potential impact on Arrhythmogenic Cardiomyopathy.

Understanding Atenolol's ATC classification helps in recognizing its therapeutic mechanisms and guiding its application in treating ACM.

**Strength and Reliability of the Link Prediction**
- Graph Explainability Metrics: The high scores for the prediction (0.9828) along with high scoring top edges indicate a robust link prediction. These metrics suggest strong associations and relevance of the connections, providing confidence in the predicted link between Atenolol (DB00335) and Arrhythmogenic Cardiomyopathy (D002313).
- Potential for Wet Lab Validation: Given the strong and biologically plausible connections, this link prediction holds promise for experimental validation. The identified pathways and disease interactions can be further investigated in wet lab settings to confirm their roles and therapeutic potential.

## Program output from running explainability analysis

```
Using CUDA: NVIDIA GPU
Summary of the Knowledge Graph:
Number of nodes: 26570
Number of edges: 3744650
Average node degree: 140.94
Number of node features: 14
Contains isolated nodes: False
Contains self-loops: False
Is undirected: True
No edge weights found.

Epoch: 1000, Loss: 0.4149
Val Accuracy: 0.7508, Test Accuracy: 0.7520
Val Precision: 0.6732, Test Precision: 0.6741
Val Recall: 0.9749, Test Recall: 0.9755
Val F1 score: 0.7964, Test F1 score: 0.7973
Val ROC AUC: 0.9634, Test ROC AUC: 0.9642
Val AUPRC: 0.9710, Test AUPRC: 0.9716
Model training successful.
Optimal prediction threshold 0.6776880025863647 which achieved f1
0.9244398181120912

Edge: (6303, 935) DrugBank_Compound:DB00571_MeSH_Disease:D002311
Predicted probability: 0.9828439950942993
('MeSH_Tree_Disease:C14.280.195.160', 'MeSH_Disease:D002311')
0.8513996005058289
('MeSH_Disease:C566467', 'MeSH_Disease:D002311') 0.8510035872459412
('MeSH_Tree_Disease:C16.320.488.750', 'MeSH_Disease:D002311')
0.8409984707832336
('MeSH_Disease:C567733', 'MeSH_Disease:D002311') 0.8351783156394958
('MeSH_Disease:C563865', 'MeSH_Disease:D002311') 0.8296864628791809
('MeSH_Disease:C567316', 'MeSH_Disease:D002311') 0.8288064002990723
('MeSH_Disease:C567407', 'MeSH_Disease:D002311') 0.8282631635665894
('MeSH_Tree_Disease:C14.280.238.070', 'MeSH_Disease:D002311')
0.824195146560669
('MeSH_Disease:C566599', 'MeSH_Disease:D002311') 0.8190982341766357
```

```
('MeSH_Disease:C567129', 'MeSH_Disease:D002311') 0.7892348170280457
('DrugBank_Compound:DB00571', 'MeSH_Disease:D002311') 1.0
Saved to file
./DrugBank_Compound:DB00571_MeSH_Disease:D002311_edge_importance.pdf

Edge: (6066, 935) DrugBank_Compound:DB00280_MeSH_Disease:D002311
Predicted probability: 0.9458465576171875
('MeSH_Disease:C566170', 'MeSH_Disease:C567507') 0.774361789226532
('MeSH_Disease:C537991', 'MeSH_Disease:C537990') 0.47969332337379456
('MeSH_Tree_Disease:C14.280.238', 'MeSH_Tree_Disease:C14.280.238.070')
0.3734593987464905
('MeSH_Disease:C564359', 'MeSH_Disease:C535493') 0.35720232129096985
('MeSH_Tree_Disease:C16.131.077.121', 'MeSH_Disease:D056889')
0.32097339630126953
('MeSH_Tree_Disease:C23.888.592.610', 'MeSH_Disease:D054971')
0.28609830141067505
('MeSH_Disease:C567507', 'MeSH_Disease:C566170') 0.25899600982666016
('MeSH_Disease:D054971', 'MeSH_Disease:D002311') 0.23935550451278687
('MeSH_Disease:C537990', 'MeSH_Disease:D002311') 0.23445141315460205
('MeSH_Tree_Disease:C14.280.238.070', 'MeSH_Disease:D002311')
0.2253691852092743
('DrugBank_Compound:DB00280', 'MeSH_Disease:D002311') 1.0
Saved to file
./explanations/DrugBank_Compound:DB00280_MeSH_Disease:D002311_edge_importance.
pdf

Edge: (6114, 937) DrugBank_Compound:DB00335_MeSH_Disease:D002313
Predicted probability: 0.9828439950942993
('MeSH_Disease:C566005', 'MeSH_Disease:C566171') 0.8301653861999512
('MeSH_Tree_Disease:C14.280.238', 'MeSH_Tree_Disease:C14.280.238.160')
0.8089691400527954
('MeSH_Disease:C566168', 'MeSH_Disease:C567654') 0.6274333596229553
('MeSH_Disease:C567505', 'MeSH_Disease:C567654') 0.564781904220581
('MeSH_Disease:C567505', 'MeSH_Disease:C566168') 0.33318817615509033
('MeSH_Disease:C563866', 'MeSH_Disease:C566005') 0.24675127863883972
('MeSH_Disease:C566171', 'MeSH_Disease:D002313') 0.24345092475414276
('MeSH_Tree_Disease:C14.280.238.160', 'MeSH_Disease:D002313')
0.2169383317232132
('MeSH_Disease:C566005', 'MeSH_Disease:D002313') 0.2045152336359024
('ATC_Class:C07BB03', 'DrugBank_Compound:DB00335') 0.19364745914936066
('DrugBank_Compound:DB00335', 'MeSH_Disease:D002313') 1.0
Saved to file
./explanations/DrugBank_Compound:DB00335_MeSH_Disease:D002313_edge_importance.
pdf
```

# Top10 Important Edges and Predicted Edge Highlighted

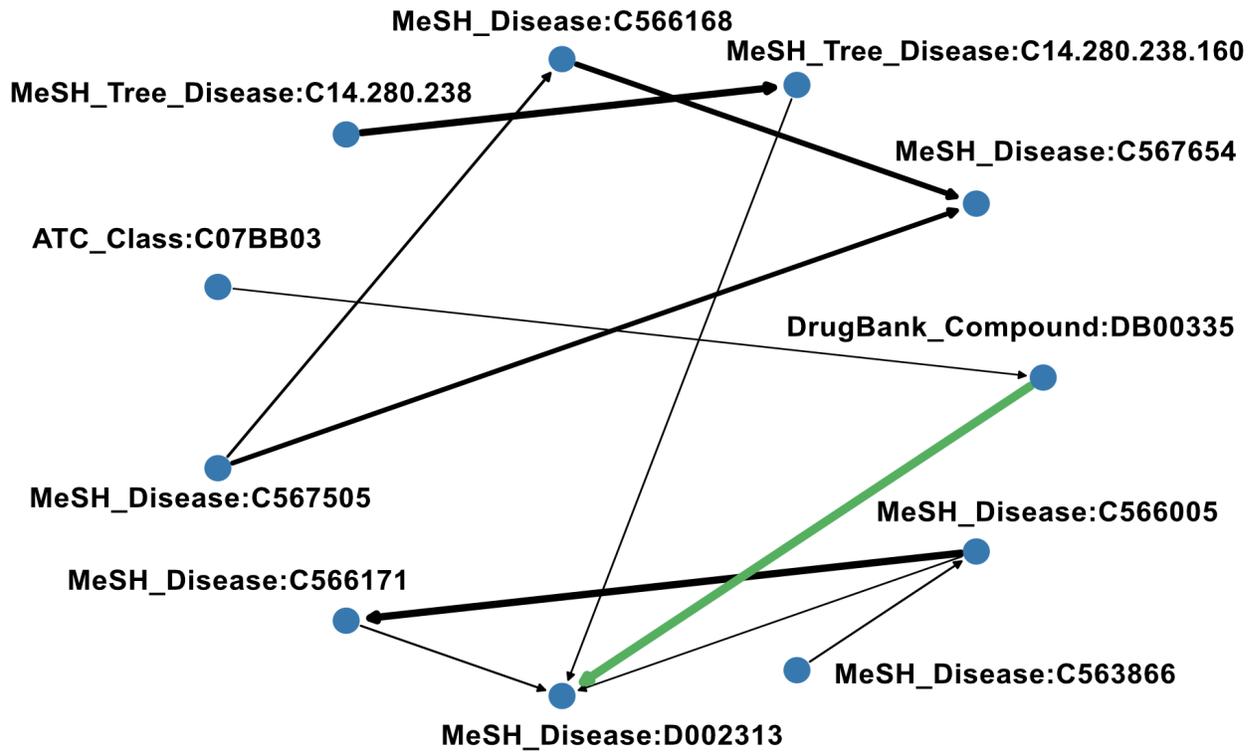

# Question 8:

## User: What literature evidence supports the claim that these predicted drugs could be used to treat ACM and/or DCM?

## RUGGED:

[Performing Literature Retrieval]

The investigation of Atenolol (DB00335) as a potential therapeutic for Arrhythmogenic Cardiomyopathy (ACM) has been supported by a combination of literature evidence. Our system identified four research publications relevant to Atenolol and ACM, including two original research articles and two clinical case reports.

**Original Research Articles**

1. Effect of beta-blockers on arrhythmias during six weeks after suspected myocardial infarction (PMID: 387170)
   - This study examined the incidence of arrhythmias in patients with suspected myocardial infarction treated with either propranolol, atenolol, or placebo. The findings indicated that beta-blockers, including atenolol, did not show significant antiarrhythmic action in the dosage used. This study raises doubts about the effectiveness of atenolol in reducing arrhythmias post-myocardial infarction, which could imply limited utility in treating ACM-related arrhythmias.
2. Comparison of the effects of xamoterol, atenolol, and propranolol on breathlessness, fatigue, and plasma electrolytes during exercise in healthy volunteers (PMID: 1685993)
   - This research investigated the effects of atenolol and other beta-blockers on heart rate, blood pressure, and electrolyte concentrations during exercise. Atenolol significantly reduced maximum exercise heart rate and blood pressure. The study suggests that atenolol's impact on exercise-induced hyperkalaemia could be beneficial in managing exercise-induced arrhythmias, which is relevant for ACM.

**Clinical Case Reports**

1. Exercise-induced ventricular arrhythmias and sudden cardiac death in a family (PMID: 9106603)
   - This case report describes a family with a history of sudden cardiac deaths and exercise-induced arrhythmias. The patients were treated with atenolol, which significantly reduced exertional arrhythmias as assessed by serial treadmill tests. This suggests atenolol may be effective in preventing exercise-induced arrhythmias, which is pertinent to ACM.
2. Sertraline-induced rhabdomyolysis in an elderly patient with dementia and comorbidities (PMID: 19567656)
   - While this report primarily discusses sertraline-induced rhabdomyolysis, it mentions atenolol as part of the patient's medication regimen without adverse interactions. This is less directly relevant but indicates atenolol's general safety profile in complex patient cases.

**Synthesis of Evidence**

The literature provides mixed support for the hypothesis that atenolol could be used to treat ACM. The original research articles offer limited evidence of atenolol's effectiveness in reducing arrhythmias associated with myocardial infarction, but suggest potential benefits in managing exercise-induced arrhythmias. The clinical case reports, particularly the one involving a family with exercise-induced ventricular arrhythmias, support atenolol's efficacy in reducing such arrhythmias, relevant to ACM.

Based on the majority of the evidence from original research and supported by clinical case reports, atenolol shows some promise in managing exercise-induced arrhythmias, which is pertinent for Arrhythmogenic Cardiomyopathy. However, its overall effectiveness and utility specifically for ACM require further validation through clinical trials and wet lab experiments.